\documentclass[]{opendatalab}
\usepackage{listings}
\usepackage{xcolor}
\usepackage{multirow}
\usepackage{hyperref}
\usepackage[T1]{fontenc}
\usepackage[utf8]{inputenc}
\usepackage{url}
\usepackage{graphicx}

\usepackage{tabularx} 
\usepackage{booktabs}
\usepackage{array}
\usepackage{colortbl}
\usepackage{xcolor}
\usepackage{wrapfig}
\usepackage{marvosym}
\definecolor{lightgreen}{RGB}{240, 255, 240}
\definecolor{lightblue}{RGB}{230, 240, 250} 

\usepackage{fvextra}   
\fvset{              
  breaklines=true,            
  breakanywhere=true,         
  breaksymbolleft=\tiny\ensuremath\hookrightarrow,
  breakindent=1.2em,         
  commandchars=\\\{\}          
}
\usepackage{xcolor}
\usepackage{longtable}
\usepackage{listings}
\usepackage[numbers]{natbib}
\usepackage{listings}
\usepackage{xcolor}

\lstdefinelanguage{json}{
    basicstyle=\small\ttfamily,
    numbers=left,
    numberstyle=\scriptsize,
    stepnumber=1,
    numbersep=8pt,
    showstringspaces=false,
    breaklines=true,
    frame=single,
    string=[s]{"}{"},
    comment=[l]{:\ "},
    morecomment=[l]{:"},
    literate=
        *{0}{{{\color{blue}0}}}{1}
         {1}{{{\color{blue}1}}}{1}
         {2}{{{\color{blue}2}}}{1}
         {3}{{{\color{blue}3}}}{1}
         {4}{{{\color{blue}4}}}{1}
         {5}{{{\color{blue}5}}}{1}
         {6}{{{\color{blue}6}}}{1}
         {7}{{{\color{blue}7}}}{1}
         {8}{{{\color{blue}8}}}{1}
         {9}{{{\color{blue}9}}}{1}
}

\definecolor{codegreen}{rgb}{0,0.6,0}
\definecolor{codegray}{rgb}{0.5,0.5,0.5}
\definecolor{codepurple}{rgb}{0.58,0,0.82}
\definecolor{backcolour}{rgb}{0.95,0.95,0.92}
\definecolor{promptcolor}{HTML}{D1D0F2}
\definecolor{promptcolorheader}{HTML}{bdbcec}
\newcommand{\promptbox}[2]{
\begin{tcolorbox}[
top=0.3em,bottom=0.3em,left=0.5em,right=0.5em,
toptitle=0.3em,bottomtitle=0.2em,boxsep=0pt,
colframe=promptcolorheader,colback=promptcolor!50,boxrule=0.5pt,
]
\footnotesize
\end{tcolorbox}
}

\title{Dripper: Token-Efficient Main HTML Extraction with a Lightweight LM}

\author[1,2\ *]{Mengjie Liu}
\author[1\ *]{Jiahui Peng}
\author[1\ *]{Wenchang Ning}
\author[1\ *]{Pei Chu}
\author[1 \ * \ \ddagger]{Jiantao Qiu}
\author[1]{Ren Ma}
\author[1,2]{He Zhu}
\author[1]{Rui Min}
\author[1]{Lindong Lu}

\author[1]{Linfeng Hou}
\author[1]{Kaiwen Liu}
\author[1]{Yuan Qu}
\author[1]{Zhenxiang Li}
\author[1]{Chao Xu}
\author[1]{Zhongying Tu}
\author[2\  \textrm{\Letter}]{Wentao Zhang}
\author[1\ \textrm{\Letter}]{Conghui He}

\affiliation[1]{Shanghai Artificial Intelligence Laboratory}
\affiliation[2]{Peking University}

\abstract{
High-quality main content extraction from web pages is a critical prerequisite for constructing large-scale training corpora. While traditional heuristic extractors are efficient, they lack the semantic reasoning required to handle the structural heterogeneity of the modern web. Conversely, well-pretrained generative Large Language Models (LLMs) offer superior document comprehension but are prohibited by excessive computational costs, limited context windows, and hallucination risks when applied at web scale. We present \textbf{Dripper}, a lightweight framework that resolves these bottlenecks through four contributions:
(1) We reformulate extraction as a \textbf{constrained sequence labeling} task using SLMs (Small Language Models). This paradigm eliminates generative hallucinations and achieves exceptional efficiency, reaching a throughput of 3.08 pages per second on a single A100 GPU.
(2) We construct \textbf{WebMainBench}, a rigorous benchmark of 7,809 human-annotated pages covering 5,434 unique domains and multiple languages. Evaluations show our Dripper-0.6B model \textbf{outperforms} heuristics like Trafilatura and rivals massive models like DeepSeek-V3.2(685B), GPT-5 and Gemini-2.5-Pro, offering an optimal efficiency-accuracy trade-off.
(3) We demonstrate infrastructural value by \textbf{pre-training a 1B model} on a Dripper-curated corpus (63B tokens). This model significantly outperforms baselines in downstream tasks, proving the critical role of extraction quality and the effectiveness of our framework.
(4) We \textbf{open-source} the Dripper-0.6B weights and codebase to facilitate the construction of high-quality datasets.
}

\metadata[* Equal contribution\quad \textrm{\Letter} Corresponding author \quad $\ddagger$ Project leader]{}
\metadata[Code:]{\url{https://github.com/opendatalab/MinerU-HTML}}
\metadata[Model:]{\url{https://huggingface.co/collections/opendatalab/mineru-html}}
\correspondence{Conghui He, \email{heconghui@pjlab.org.cn}; Wentao Zhang, \email{wentao.zhang@pku.edu.cn}}

\begin{document}

\maketitle

\section{Introduction}

The World Wide Web forms the foundational data repository for modern AI, serving as the primary source for training corpora like C4~\cite{C4} and for building the knowledge graphs that power large-scale applications~\cite{applications}. The scale of this resource is immense, with archiving initiatives like Common Crawl~\cite{commoncrawl} ingesting billions of new pages monthly. However, unlocking the value of this massive volume presents a formidable barrier: raw, unstructured HTML must be efficiently converted into high-quality, structured data. Consequently, the development of computationally efficient and scalable content extraction methodologies has become a critical prerequisite for the entire downstream information processing pipeline~\cite{web2text}.

The primary bottleneck lies in the inability of traditional extraction paradigms to cope with the web's inherent heterogeneity. Although HTML standards define semantic tags (e.g., <article>), their practical adoption is notoriously inconsistent~\cite{wang2022webformer}. Modern web pages frequently prioritize visual layout over semantic structure, often fragmenting the main text with interspersed boilerplate (e.g., in-feed advertisements) or encapsulating critical data-such as code blocks and tables-within generic <div> tags. This structural ambiguity renders rule-based parsers unreliable, leading to both the truncation of body text and the corruption of high-value structured knowledge. Similarly, heuristic methods relying on statistical priors-such as text-to-tag ratios-often falter, as even pages sharing the same template can exhibit vast statistical variance due to content differences. Furthermore, vision-based approaches~\cite{diffbot2025extract} are largely inapplicable to large-scale offline archives like Common Crawl, which typically store raw HTML without the CSS stylesheets requisite for visual rendering. These structural and operational constraints prevent established heuristic and vision-based methods from achieving the generalization required for web-scale processing. 

The semantic reasoning capabilities of LLMs offer a promising theoretical solution for parsing complex web structures~\cite{wang2025readerlm}. However, their direct application is thwarted by three severe practical barriers. First, \textbf{excessive context length} renders processing raw HTML computationally infeasible at scale. Our analysis of 14,000 Common Crawl files reveals that 29.3\% of pages exceed 32k tokens and 21.0\% surpass 128k tokens-lengths that overwhelm the context windows of efficient LLMs. Second, the \textbf{structural complexity} of HTML presents a critical dilemma. While stripping tags effectively reduces input length, it simultaneously discards vital structural cues necessary for distinguishing main content from boilerplate. Finally,  generative models are prone to \textbf{fabricating content} not present in the source~\cite{ji2023survey}, a critical failure mode for extraction tasks demanding strict fidelity.

To surmount these challenges, we introduce \textbf{Dripper}, a framework that reframes content extraction as a token-efficient sequential block classification task. Central to our design is a dual-branch representation strategy that decouples semantic reasoning from content reconstruction. Specifically, the framework transforms raw input into two parallel, block-aligned sequences: \textbf{Simplified HTML} and \textbf{Mapping HTML}. The former undergoes aggressive pruning to minimize token count, enabling a lightweight 0.6B parameter model, Dripper-0.6B, to efficiently analyze the document structure within a compressed context window. Crucially, a constrained decoding mechanism enforces structured output, simultaneously eliminating hallucinations and accelerating inference. These high-confidence labels are then applied to the corresponding blocks in the \textbf{Mapping HTML}-which retains the original DOM fidelity-to reconstruct the final output. This design effectively harmonizes efficiency with fidelity, enabling high-speed processing while preserving the extracted content as a precise, stylistically intact DOM subtree.

Our main contributions are summarized as follows:

(1) We propose \textbf{a novel extraction paradigm} using SLMs, redefining the task as constrained sequence labeling. By transforming raw HTML into simplified block sequences and enforcing a fixed label output space, we systematically eliminate generative hallucinations and achieve a processing speed of 3.08 pages per second on a single A100 GPU.

(2) We construct \textbf{WebMainBench}, a high-precision dataset comprising over 7,800 web pages sampled from Common Crawl, spanning 5,434 unique domains and multiple languages. Validated through a rigorous multi-round human annotation protocol, it establishes the most reliable standard to date for unified evaluation. Extensive experiments demonstrate that our Dripper-0.6B model(F1: \textbf{0.8779}) not only significantly outperforms traditional heuristics like Trafilatura (F1: \textbf{0.6402}) and Resiliparse (F1: \textbf{0.6290}), but also approaches the performance of massive commercial models like DeepSeek-V3.2 (F1: \textbf{0.9098}) , GPT-5 (F1: \textbf{0.9024}) and Gemini-2.5-Pro (F1: \textbf{0.8979}), representing an optimal trade-off between efficiency and accuracy.

(3) We empirically demonstrate the critical role of extraction quality in foundation model training. By leveraging Dripper to \textbf{curate a web corpus of approximately 63 billion tokens for pre-training a 1B parameter model}, we observe significant performance gains on downstream tasks compared to baselines trained on data from other tools, highlighting the infrastructural value of our framework in the modern data ecosystem.

(4) We publicly release the \textbf{Dripper-0.6B} model weights alongside the complete codebase, including the full extraction pipeline and evaluation scripts. This release aims to facilitate reproducibility and accelerate community efforts in constructing higher-quality corpora. Our trained model weights, code and the WebMainBench benchmark are publicly available.

\section{Related Work}
Main text extraction aims to extract main content from raw HTML while filtering out boilerplate elements such as navigation and advertisements, a critical technique for building high-quality web corpora. The methods for accomplishing this task have evolved through several distinct paradigms, each addressing the limitations of its predecessor.

\textbf{Heuristic and rule-based Methods}. Early approaches predominantly relied on manually engineered heuristics to distinguish main content from boilerplate. These methods operate on the observation that content-rich regions differ structurally from noisy elements, using features like text-to-tag ratios (CETR)~\cite{weninger2010cetr}, visual cues from the rendered page (VIPS)~\cite{cai2003vips}, or a combination of heuristics such as link and stop-word density (Readability~\cite{MozillaReadabilityJS}, jusText~\cite{pomikalek2011removing}). While computationally efficient, these methods are often brittle and require continuous maintenance to adapt to evolving web design patterns.

\textbf{Supervised Learning Methods}. To move beyond handcrafted rules, subsequent work approached body text extraction as a supervised machine learning problem. This paradigm shift began with classic methods like Boilerpipe~\cite{kohlschutter2010boilerplate}, Dragnet~\cite{peters2013content}, which treated the task as a classification problem using manually designed features. The advent of deep learning marked a further evolution from feature engineering to representation learning. ~\cite{vogels2018web2text,leonhardt2020boilerplate,zhou2021simplified}. To better leverage the hierarchical structure of HTML, subsequent research introduced Graph Neural Networks (GNNs)~\cite{zhou2021simplified} and Transformer-based architectures like WebFormer~\cite{wang2022webformer}, which improved extraction accuracy by capturing complex relationships between nodes. While achieving higher accuracy, these models often require substantial labeled data, and their complex architectures incur significant computational overhead.

\textbf{Hybrid Systems and Production Tools}. In parallel with academic advancements, a suite of powerful open-source tools has emerged, often blending multiple techniques for practical application. Trafilatura~\cite{barbaresi2021trafilatura}  has become a strong baseline by integrating a sophisticated cascade of rules with established algorithms like jusText~\cite{pomikalek2011removing} and Readability~\cite{MozillaReadabilityJS} as fallbacks. Other tools, such as magic-html~\cite{magichtml}, focus on simplifying complex HTML structures before extraction, often as part of larger document AI ecosystems. More recently, frameworks such as crawl4ai~\cite{crawl4ai2024} have adopted an explicitly hybrid architecture, combining rule-based selectors, traditional machine learning, and LLMs to provide versatile solutions for AI data pipelines.

\textbf{Generative-Language-based Methods}. Recent months have seen rapid progress in decoder-only large language models. Base models pre-trained on massive, high-quality, and highly-diverse corpora have become the de-facto starting point for most NLP tasks. The most representative work in this line is ReaderLM-v2~\cite{wang2025readerlm}, which frames main-content extraction as an HTML-to-Markdown translation problem. Starting from a 1.5 B-parameter Qwen2.5 checkpoint, the authors first extend the context window to 512 k tokens through continual pre-training, then fine-tune with supervised fine-tuning (SFT) and direct-preference optimization (DPO) to produce clean Markdown. This pipeline reuses the open-source model zoo and inference-acceleration stacks already available in the LLM community. Nevertheless, even the official best-practice implementation \footnote{https://huggingface.co/jinaai/ReaderLM-v2} still expects the full, un-pruned HTML page as input and generates the complete body text in one pass. This incurs heavy computational overhead and, during long-sequence generation, often produces unwanted artifacts such as repetitions or unescaped HTML tags. Consequently, the potential of SLMs for extraction remains largely untapped.

\section{Methodology}
\label{Methodology}

In this section, we elaborate on the technical architecture of the Dripper framework. We begin with a overview of the system's dual-branch strategy in §\ref{sec:overview}. Subsequently, §\ref{sec:preprocessing} details the processing pipeline that transforms raw HTML into parallel simplified and mapping representations. In §\ref{sec:model}, we introduce the Dripper-0.6B model, discussing its training paradigm and the constrained decoding mechanism employed to guarantee output validity. Finally, §\ref{sec:formulation} provides the formal mathematical formulation of the task.

\subsection{System Architecture Overview}
\label{sec:overview}

\begin{figure}
  \includegraphics[width=\textwidth]{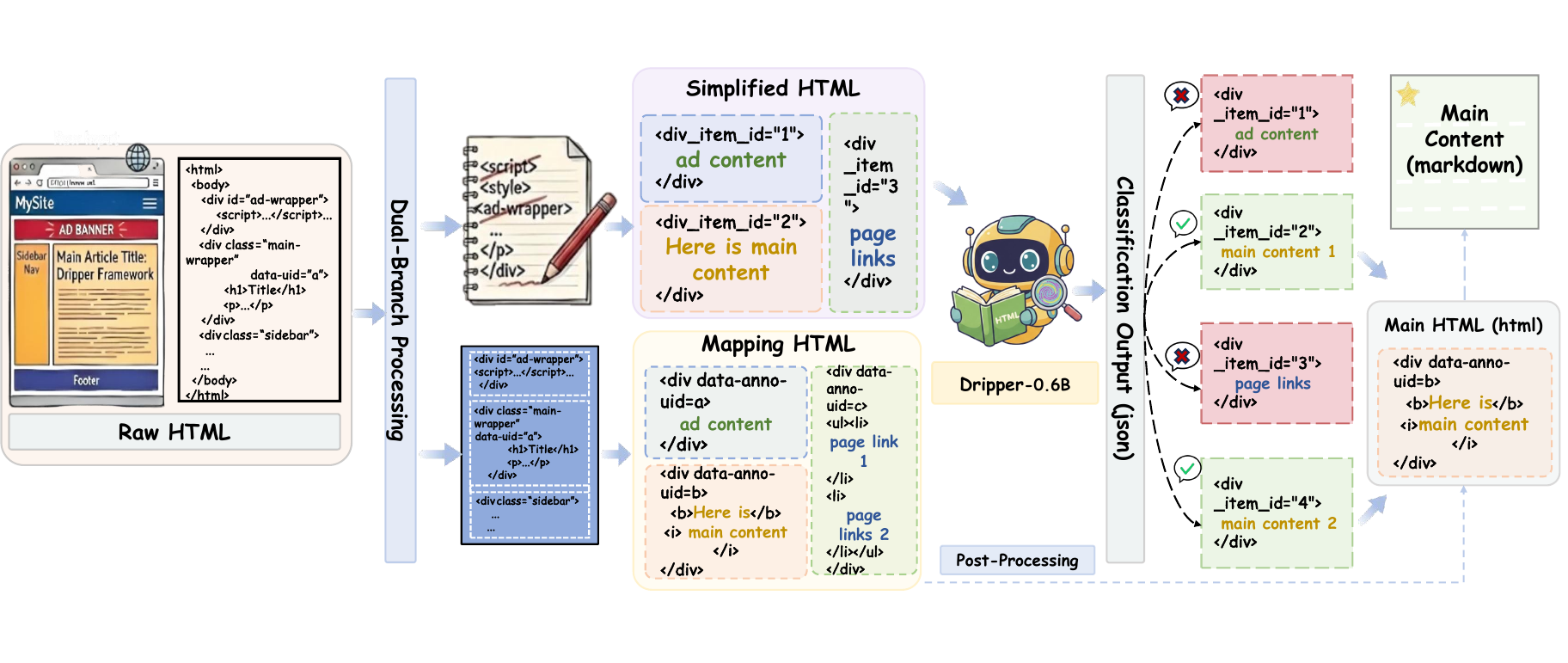}
  \caption{Overview of the Dripper framework. The system balances efficiency and fidelity via a three-stage pipeline. (1) Pre-processing: Raw HTML is converted into parallel, aligned sequences---Simplified HTML for efficient model input and Mapping HTML for faithful reconstruction. (2) Extraction: Dripper-0.6B predicts semantic labels on the simplified sequence, governed by a constrained decoding mechanism. (3) Post-processing: Predicted labels are applied to the Mapping HTML to reconstruct high-fidelity content and generate the final clean Markdown.}
  \label{fig:teaser}
\end{figure}

The Dripper framework balances computational efficiency with extraction fidelity through a three-stage pipeline centered on a dual-branch representation strategy.

In the first stage (Pre-processing), the raw HTML is transformed into two parallel, block-aligned sequences: \textbf{Simplified HTML} and \textbf{Mapping HTML}. This bifurcation allows us to decouple semantic reasoning from content reconstruction. In the second stage (Extraction), \textbf{Dripper-0.6B} ingests only the token-efficient Simplified HTML to predict semantic labels, utilizing a constrained decoding mechanism to ensure validity. Finally, in the third stage (Post-processing), these labels are applied back to the Mapping HTML. This step retrieves the corresponding high-fidelity content blocks to construct the final output, thereby recombining the efficiency of the simplified branch with the structural integrity of the mapping branch.

\subsection{Dual-Branch Generation and Reconstruction}
\label{sec:preprocessing}

Raw HTML is inherently verbose to support precise visual rendering, necessitating complex nested tags and attributes that act as noise for semantic interpretation. Naively feeding this data into SLMs results in prohibitive token costs and context overflows. To resolve this, we decouple semantic reasoning from content reconstruction. We employ a \textbf{dual-branch strategy} that generates two parallel representations bridged by shared segmentation logic: a Simplified HTML branch, which aggressively prunes visual noise to reduce token count by approximately \textbf{84.4\%} for efficient inference; and a Mapping HTML branch, which retains full DOM fidelity for precise content retrieval.

For the Simplified HTML, which serves as the input for the SLMs, we apply a four-step distillation process: (1) preemptively removing non-content structural tags such as <style>, <script>, <header> and <aside>; (2) pruning all attributes except for class and id, which often carry the most valuable semantic cues for distinguishing content blocks; (3) aggregating tags into semantic blocks based on heuristic rules, specifically targeting elements that induce line breaks while treating cohesive units like tables <table> and lists <ul> as indivisible to preserve their integrity.

The Mapping HTML undergoes the identical (1) DOM Cleaning and (2) Attribute Pruning processes. This redundancy is critical for two reasons: it ensures strict node-level alignment with the Simplified branch (preventing index mismatch during label projection) and guarantees that the final extracted content is free from operational noise (e.g., scripts and inline styles).

The bifurcation occurs at step (3) Block Aggregation. While the Simplified HTML aggregates adjacent inline tags into coarse-grained tokens to compress sequence length, the Mapping HTML explicitly bypasses this step, preserving the precise subtree structure of every distinct block.

Consequently, the Simplified HTML is ingested by the model for efficient label prediction, while the Mapping HTML is utilized in the post-processing stage for high-fidelity content retrieval. This decoupling strategy is highly effective: it reduces the mean input sequence length \textbf{from 44,590 tokens (Raw) to 6,960 tokens (Simplified)}-a dramatic 84.4\% reduction-thereby enabling the 0.6B model to reason over long documents within a standard context window, while guaranteeing that the final extraction remains a precise, structurally intact DOM subtree.

\subsection{Sequential Block Classification Model}
\label{sec:model}

We instantiate our extraction engine using Qwen3-0.6B~\cite{qwen3technicalreport}, the most compact variant in the Qwen3 series. This model is selected for its optimal balance between parameter efficiency and semantic capability, featuring a 32K context window and robust support for over 100 languages.

To align the base model with our extraction paradigm, we employ Supervised Fine-Tuning (SFT) using the Llama-Factory framework~\citep{zheng2024llamafactory}. The model is fine-tuned on a large-scale corpus comprising approximately 986k diverse web samples, the construction of which is detailed in Section~\ref{sec:training_data}. The training process spans a fixed duration of 4 epochs. We select the final checkpoint as the production model, denoted as \textbf{Dripper-0.6B}. Detailed hyperparameter settings and training configurations are provided in Appendix~\ref{app:training_config}.

Specifically, Dripper-0.6B is trained to generate predictions following a strict JSON schema, mapping block IDs to binary labels (e.g., \{"1": "main", "2": "other"\}). Leveraging this deterministic format, we employ \textbf{regex-based constrained decoding} during inference to strictly govern the generation process. This mechanism offloads syntactic compliance to the inference engine, enabling the model to dedicate its limited capacity entirely to semantic discrimination. Crucially, it not only eliminates format hallucinations but also prevents the repetitive generation loops often prone to SLMs, thereby guaranteeing consistent inference efficiency.

\subsection{Task Formulation}
\label{sec:formulation}

The system architecture detailed above effectively transforms the
content extraction task into a well-defined sequence labeling
problem. Formally, the proposed framework transforms main content extraction into a sequence labeling problem. Let the pre-processed document be represented as a sequence of $n$ simplified blocks, denoted as $X = [x_1, x_2, \dots, x_n]$. Each block $x_i$ corresponds to a ground-truth binary label $y_i \in \{0, 1\}$, where $1$ represents main content and $0$ represents boilerplate.

The objective is to learn a mapping function $f_\theta$ (parameterized by Dripper-0.6B) that predicts the label sequence $\hat{Y} = [\hat{y}_1, \hat{y}_2, \dots, \hat{y}_n]$ conditioned on the input $X$:
\begin{equation}
\hat{Y} = \operatorname*{argmax}_{Y} P(Y \mid X; \theta)
\end{equation}
subject to the constraint that $Y$ must conform to the valid JSON format enforced by the decoding constraints.

Finally, the extracted content $C$ is reconstructed by applying the predicted mask $\hat{Y}$ to the parallel Mapping HTML sequence $M = [m_1, m_2, \dots, m_n]$. Since $X$ and $M$ are strictly aligned (i.e., $x_i$ and $m_i$ refer to the same semantic unit), the final output is obtained as:
\begin{equation}
C = \{ m_i \in M \mid \hat{y}_i = 1 \}
\end{equation}
This formulation ensures that the output $C$ is a precise subset of the original DOM tree, preserving all structural and stylistic fidelity.

\section{Dataset and Benchmark}
In this section, we detail the construction of our large-scale training dataset (Section~\ref{sec:training_data}) and our new evaluation benchmark, WebMainBench (Section~\ref{sec:webmainbench}), along with its evaluation metrics.

\subsection{Training Data Construction}
\label{sec:training_data}
To ensure robust model generalization, we curate a large-scale, high-diversity training dataset through a rigorous three-stage pipeline designed to maximize coverage across page layouts, languages, and document formats.

\textbf{Stage 1: Layout-Aware Sampling.} To capture structural heterogeneity, we aggregate pages by domain across 107 Common Crawl snapshots. For each domain, we extract structural features from the DOM trees (capping large domains at 10,000 randomly sampled pages) and compute pairwise cosine similarities. We then apply DBSCAN clustering to these feature vectors to identify distinct layout patterns. From each of the resulting approximately 40 million unique clusters, we sample a single representative webpage, yielding a candidate pool of \textbf{40 million} structurally diverse pages.

\textbf{Stage 2: Multilingual and Format Filtering.} We refine this candidate pool to ensure linguistic balance and format diversity. First, we extract the main text using Trafilatura and employ FastText lid-176\footnote{https://fasttext.cc/docs/en/language-identification.html} for language identification. This process produces a balanced subset of 10 million pages, comprising 4.75M English, 4.75M Chinese, and 0.5M pages in other languages. To prevent format bias, we further categorize these pages using a specialized web format classifier~\citep{organize_the_web_citation_key}. We then perform stratified sampling across the identified formats, resulting in a curated set of approximately 1 million pages (485k English, 487k Chinese, 50k others) for annotation.

\textbf{Stage 3: Automated Annotation.} The selected candidates are processed via our HTML simplification algorithm (see §\ref{sec:preprocessing}) to generate input sequences. We then utilize the \textbf{DeepSeek-V3.2} API to generate block-level labels for the Simplified HTML using a strictly defined prompt (see Appendix Figure~\ref{code:prompt}). Following a quality control step that discards samples yielding no main content (i.e., where all blocks are labeled as 'other'), we establish a final training corpus of \textbf{986,442 samples}.

\subsection{WebMainBench: A Modern High-Precision Benchmark}
\label{sec:webmainbench}

Existing benchmarks often suffer from limited scale, archaic web patterns, or inconsistent annotation standards. To address these gaps, we present \textbf{WebMainBench}, a rigorously annotated dataset comprising \textbf{7,809 samples}. For detailed construction protocols and the definitions of our difficulty stratification (Simple, Mid, Hard), please refer to Appendix~\ref{appdx:benchmark}.

\subsubsection{Composition and Statistics}
We employ a hybrid sampling strategy to ensure breadth and quality: 90\% of pages are randomly drawn from Common Crawl to capture "long-tail" diversity, while 10\% are sampled from high-ranking sites (via Chinaz Alexa) to cover professional layouts. The dataset demonstrates high entropy, spanning \textbf{5,434 unique domains} across \textbf{150 Top-Level Domains (TLDs)} and \textbf{46 languages}. This broad distribution ensures the benchmark is not skewed by dominant sources. Furthermore, semantic classification via GPT-5 confirms coverage of diverse page types, ranging from standard news articles to complex forums (see Appendix~\ref{app:benchmark_stats}).

\subsubsection{Superiority over Legacy Benchmarks}
While we include the aggregated \textbf{Web Content Extraction Benchmark (MCEB)} for fair comparison, WebMainBench is explicitly developed to resolve three fundamental limitations of legacy collections:

First, unlike legacy datasets that provide only plain text, WebMainBench is annotated at the \textbf{HTML tag level} using a custom-built tool (Appendix~\ref{app:annotation_tool}). This structural fidelity serves as a unified "source of truth," enabling precise boundary evaluation and deterministic conversion into any target format (e.g., Markdown or Text), thereby eliminating ambiguity.

Second, MCEB aggregates older datasets with varying standards and frequent artifacts~\citep{survey}. In contrast, WebMainBench is a native dataset built on a unified pipeline. We enforce strict annotation rules governed by \textit{Contextual Integrity} (grouping integral content like abstracts) and \textit{Human-Generated Content} (prioritizing substance over metadata). This standardization ensures the benchmark faithfully reflects the complexity of the modern web.

Third, we enrich every sample with \textbf{multidimensional metadata}, including quantified difficulty levels, style tags, and flags for rich content (e.g., tables, LaTeX equations). This metadata layer supports the fine-grained diagnostic analysis presented in Section~\ref{sec:experiments}.

\subsubsection{Evaluation Metrics}
To standardize evaluation across tools with varying output capabilities, all extracted content is converted to a canonical Markdown format using html2text. We report the \textbf{ROUGE-N F1} score (N=5, tokenized by jieba) as the primary metric instead of ROUGE-L, chosen for its efficiency in measuring content overlap on long documents.

\section{Experiments}
\label{sec:experiments}

\subsection{Experimental Setup}

\textbf{Baseline Methods.} To comprehensively evaluate Dripper, we benchmark it against a wide spectrum of established and state-of-the-art systems. These include classic heuristic/rule-based parsers , supervised learning approaches, and recent LLM based extractors. A detailed characterization of each baseline is provided in Appendix Table~\ref{tab:baselines}.

\textbf{Evaluation Protocol.} To ensure fair comparison across tools with diverse output formats, we standardize evaluation modes: \textbf{-HTML+MD} for tools generating intermediate HTML converted to Markdown; \textbf{-MD} for native Markdown outputs; and \textbf{-TEXT} for native plain text. 

\textbf{Handling Context Limitations.} Since Dripper operates within a fixed context window (32k tokens), inputs exceeding this limit are assigned a score of 0 to reflect the system's limitation strictly. However, to assess practical applicability, we also introduce \textbf{Dripper\_fallback}. Following the design of robust crawlers like Trafilatura, this variant invokes a heuristic fallback (specifically Trafilatura) solely when Dripper encounters oversized inputs or returns empty results, allowing us to evaluate the system's viability in production environments.

\subsection{Computational Efficiency Analysis}
\label{sec:efficiency}

The computational cost of a decoder-only language model is dominated by sequence length. The inference complexity can be approximated by Equation (\ref{equation}):

\begin{equation}
\label{equation}
\text{Cost} \approx 4 L d \left( N^2 + M N + \frac{M^2}{2} \right) + 24 L d^2 (N + M) \quad \text{FLOPs}
\end{equation}

where $L$ and $d$ denote the number of layers and hidden-state dimension ($L=28, d=1024$ for Qwen3-0.6B), while $N$ and $M$ represent input and output token counts, respectively. The detailed derivation of this estimation is provided in Appendix~\ref{app:cost_estimation}.

\begin{table}[h!]
\centering
\small
\caption{Computational overhead comparison on WebMainBench. "Without" denotes the naive baseline (Raw HTML $\to$ Markdown), while "With" denotes the Dripper framework (Simplified HTML $\to$ JSON Labels). We report mean/median token lengths and estimated FLOPs. The "Ratio" row highlights the significant efficiency gains achieved by Dripper.}
\renewcommand{\arraystretch}{1.2}

\resizebox{\linewidth}{!}{%
    \begin{tabular}{c|cc|cc|cc}
    \toprule
    \multirow{2}{*}{Pre-process}
     & \multicolumn{2}{c|}{Input length (tokens)}
     & \multicolumn{2}{c|}{Output length (tokens)}
     & \multicolumn{2}{c}{Cost estimate (FLOPs)} \\
     & mean & median & mean & median & mean & median \\
    \midrule
    Without & 44590.7 & 31890 & 2345.3 & 687 & $4.630\times 10^{14}$ & $1.471\times 10^{14}$ \\
    With    & 6960.7  & 4320  & 401.4  & 213 & $2.046\times 10^{13}$ & $5.452\times 10^{12}$ \\
    \midrule
    Ratio   & 15.61\% & 13.55\% & 17.12\% & 31.00\% & 4.42\% & 3.71\% \\
    \bottomrule
    \end{tabular}%
}

\label{tab:cost}
\end{table}

To quantify efficiency gains, we compare Dripper against a naive generative baseline that inputs Raw HTML and generates full Markdown. As shown in Table~\ref{tab:cost}, our approach achieves efficiency through a dual-compression mechanism. First, the pre-processing pipeline aggressively filters noise, reducing the mean input length ($N$) to just \textbf{15.61\%} of the raw HTML. Second, by reformulating the extraction as a classification task, the output length ($M$) is compressed to \textbf{17.12\%} of the full content. These synergistic reductions lead to a dramatic decrease in computational load, lowering the mean inference cost to merely \textbf{4.42\%} of the naive baseline. This makes SLM-based extraction not only accurate but also computationally feasible for high-throughput pipelines.

\begin{figure}[t!]
  \centering
  \includegraphics[width=\linewidth]{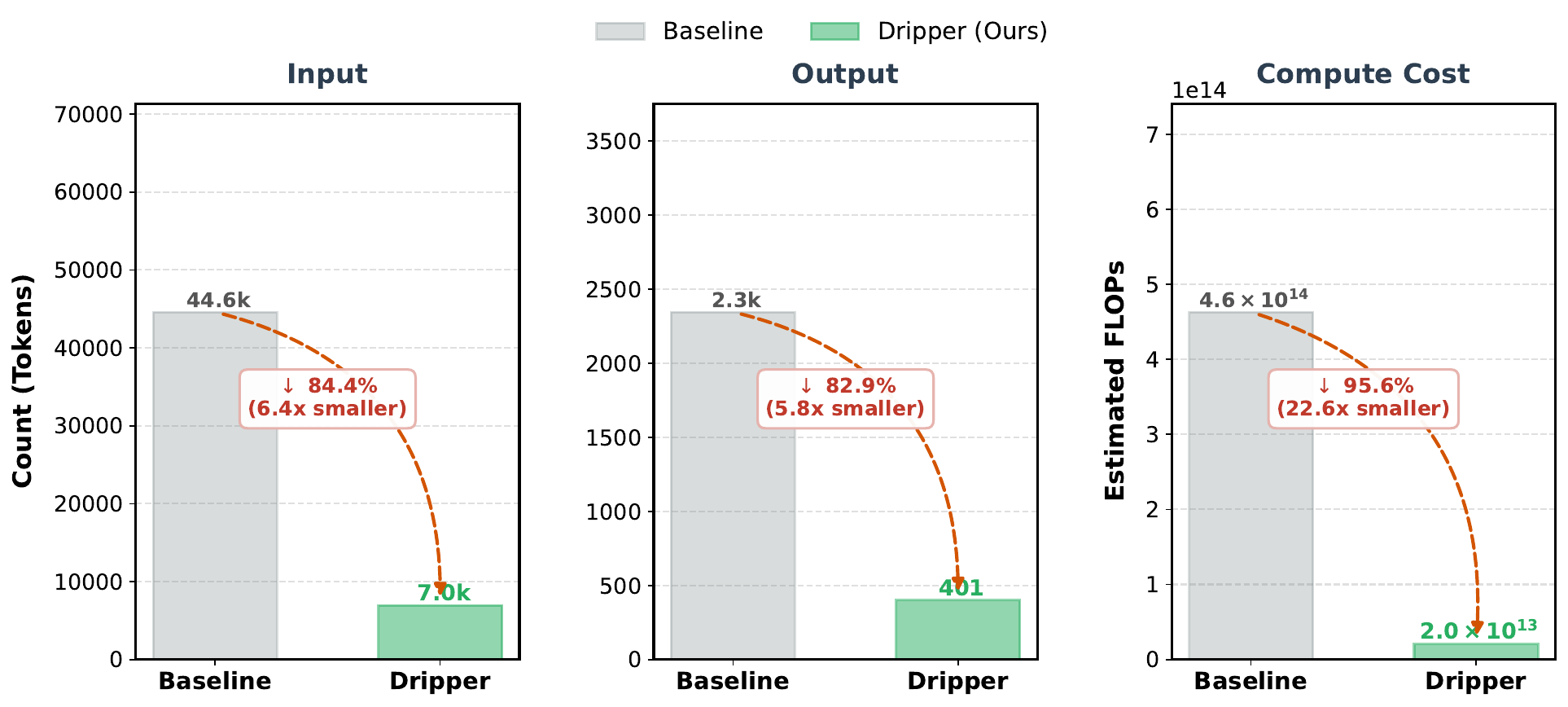}
  
  \caption{\textbf{Efficiency comparison between Baseline and Dripper.} The charts visualize the substantial reduction in resource usage across three dimensions: \textbf{Mean Input Tokens} (left), \textbf{Estimated Compute Cost} in FLOPs (center), and \textbf{Mean Output Tokens} (right). By simplifying the input HTML, Dripper reduces the average input length by $6.4\times$ (84.4\%) and, combined with constrained decoding, lowers the total computational overhead by a factor of $22.6\times$ (95.6\%) compared to the raw HTML baseline.}
  
  \label{fig:efficiency_comparison}
\end{figure}

\subsection{Results on WebMainBench}
\label{sec:main_results}

We present the comprehensive performance comparison on WebMainBench in Table~\ref{tab:main_results}. The quantitative results indicate that Dripper achieves superior performance compared to traditional extractors, establishing a new state-of-the-art among local execution methods. Specifically, the standalone Dripper model attains an overall F1 score of \textbf{0.8779}, surpassing the strongest heuristic baseline, magic-html (0.7138), by a substantial margin. This performance gap is even more distinct when compared to widely used industry standards like Trafilatura (0.6402).

\textbf{Performance on Complex Structures.} A deeper analysis reveals that our approach excels in scenarios with complex layouts. On pages categorized as 'Hard' difficulty, Dripper maintains a robust F1 score of 0.8313, whereas \texttt{magic-html} drops to 0.6434. This resilience is particularly evident in the extraction of conversational content (e.g., forums), where Dripper achieves \textbf{0.8493}, significantly outperforming heuristic parsers that typically fail to separate dialogue from boilerplate (e.g., Trafilatura at 0.5769).

\textbf{Comparison with Proprietary Frontier Models.} To establish a theoretical upper bound for extraction performance, we evaluate leading commercial Large Language Models, including DeepSeek-V3.2, GPT-5, and Gemini-2.5-Pro. Crucially, direct processing of raw HTML is infeasible even for these giants due to prohibitive token costs and context limits. Therefore, we integrate these models as drop-in replacements within the Dripper pipeline: they ingest the \textit{Simplified HTML} to predict block labels, which are then reconstructed using our standard post-processing module.

As shown in the Frontier Models section of Table~\ref{tab:main_results}, these models achieve F1 scores in the range of 0.89--0.91. Remarkably, despite having only 0.6B parameters, \textbf{Dripper\_fallback} (0.8925) delivers performance highly competitive with these commercial giants. For instance, the performance gap between our method and GPT-5 (0.9024) is \textbf{less than 1 percentage point}. This result highlights the efficacy of our task reformulation: by simplifying the input and identifying structural patterns, a specialized Small Language Model can approximate the capability of general-purpose frontier models while offering orders of magnitude lower latency and cost.

\begin{table*}[h]
\centering
\small
\caption{{Performance comparison on WebMainBench (ROUGE-N F1). Methods are grouped into Standard Extraction, Commercial LLMs , and Our Method. Results are stratified by Overall performance, Difficulty Level, and Rich Content Types.}}
\resizebox{\linewidth}{!}{%
\begin{tabular}{llccccccccc}
\toprule
Name                & Mode &   All    &   Simple    &   Mid    &   Hard      &    Table    &   Code    &   Equation      &   Conversation    \\
\midrule
\textit{Baselines} & & & & & & & & & \\
magic-html~\citep{magichtml}         & Html+MD &   0.7138 &      0.7857 &   0.7121 &    0.6434  &      0.6724 &          0.8561 &              0.8503 &           0.4688 \\
Readability~\citep{MozillaReadabilityJS}        & Html+MD &   0.6543 &      0.7415 &   0.6550 &    0.5652  &      0.5949 &          0.7803 &              0.7831 &           0.4646 \\
Trafilatura~\citep{barbaresi2021trafilatura}        & Html+MD &   0.6402 &      0.7309 &   0.6417 &    0.5466  &      0.5544 &          0.6044 &              0.7356 &           0.5769 \\
{Resiliparse \citep{bevendorff:2018}} & {TEXT} & {0.6290} & {0.7140} & {0.6323} & {0.5388} & {0.5529} & {0.6551} & {0.7859} & {0.5365} \\
Trafilatura        &   MD    &   0.6280 &      0.7148 &   0.6306 &    0.5368  &      0.5436 &          0.5778 &              0.7196 &           0.5774 \\
Trafilatura        &  TEXT   &   0.6090 &      0.6929 &   0.6114 &    0.5211  &      0.5308 &          0.5602 &              0.6983 &           0.5670 \\
html2text~\citep{html2text_pypi}           & Html+MD &   0.6042 &      0.7535 &   0.5858 &    0.4781  &      0.6000 &          0.7797 &              0.7156 &           0.5510 \\
BoilerPy3~\citep{boilerpy3}          &  TEXT   &   0.5435 &      0.6372 &   0.5466 &    0.4445  &      0.4401 &          0.4868 &              0.6616 &           0.4722 \\
GNE~\citep{gne}                 & Html+MD &   0.5172 &      0.6502 &   0.4945 &    0.4133  &      0.4158 &          0.5540 &              0.6184 &           0.3298 \\
news-please~\citep{Hamborg2017}         &  TEXT   &   0.5032 &      0.5413 &   0.5253 &    0.4351  &      0.4225 &          0.5145 &              0.6727 &           0.4054 \\
BoilerPy3          & Html+MD &   0.4803 &      0.6475 &   0.4729 &    0.3213  &      0.3818 &          0.5599 &              0.6180 &           0.4127 \\
jusText~\citep{pomikalek2011removing}            &  TEXT   &   0.4783 &      0.5158 &   0.5077 &    0.4008  &      0.3974 &          0.3810 &              0.6678 &           0.5228 \\
Goose3~\citep{goose3}              &  TEXT   &   0.4372 &      0.4534 &   0.4657 &    0.3824  &      0.3615 &          0.2915 &              0.6404 &           0.3084 \\
ReaderLM-v2~\citep{wang2025readerlm}            &   TEXT  &   0.2280 &      0.3387 &   0.2091 &    0.1414  &      0.1823 &          0.2445 &              0.2937 &           0.1510 \\
\midrule
\textit{Frontier Models} & & & & & & & & & \\
DeepSeek-V3.2      &   Html+MD  &   0.9098 &      0.9415 &   0.9104 &    0.8771  &      0.9096 &          0.9110 &              0.9286 &           0.9500 \\
GPT-5              &   Html+MD  &   0.9024 &      0.9382 &   0.9042 &    0.8638  &      0.9016 &          0.9068 &              0.9161 &           0.9432 \\
Gemini-2.5-Pro     &   Html+MD  &   0.8979 &      0.9345 &   0.8978 &    0.8610  &      0.8985 &          0.8944 &              0.9090 &           0.9452 \\
\midrule
\textit{Ours} & & & & & & & & & \\
\rowcolor{lightgreen}
Dripper & Html+MD & \textbf{0.8779} & \textbf{0.9205} & \textbf{0.8804} & \textbf{0.8313} & \textbf{0.8527} & \textbf{0.8926} & \textbf{0.9296} & \textbf{0.8537} \\
\rowcolor{lightgreen}
Dripper\_fallback  & Html+MD &   \textbf{0.8925} &      \textbf{0.9325} &   \textbf{0.8958} &    \textbf{0.8477}  &      \textbf{0.8685} &          \textbf{0.9111} &              \textbf{0.9444} &           \textbf{0.8866} \\
\bottomrule
\end{tabular}
}
\label{tab:main_results}
\end{table*}

\subsection{Generalization on Legacy Benchmarks}
\label{sec:wceb_results}

To assess the generalization capabilities of Dripper beyond our native dataset, we evaluated it on the Web Content Extraction Benchmark (WCEB), a consolidated dataset designed to address the inconsistencies of legacy collections. Unlike WebMainBench, WCEB provides ground truth primarily in plain text. Consequently, we adapted our evaluation protocol to the Html+TEXT mode, utilizing the html-text\footnote{https://pypi.org/project/html-text/} library to convert our structural output into comparable plain text.

The results, summarized in Table~\ref{tab:wceb_results}, confirm the robustness of our approach. Despite the domain shift and the loss of structural granularity in the evaluation metric, the standalone Dripper model achieves an F1 score of \textbf{0.8156}, surpassing the previous state-of-the-art heuristic method, Trafilatura (0.7832). This indicates that the semantic representations learned by Dripper are not merely artifacts of our specific annotation schema but reflect a generalized understanding of web content boundaries. Furthermore, consistent with our findings on WebMainBench, the Dripper\_fallback strategy further boosts performance to \textbf{0.8317}. This performance consistency across both modern, structure-rich benchmarks and legacy, text-centric datasets demonstrates that Dripper effectively bridges the gap between precision and generalization.

\begin{table}[h!]
\centering
\caption{{Generalization performance on WCEB (ROUGE-N F1). Since WCEB lacks rich content metadata, results are stratified only by Overall performance and Difficulty Level. The "Html+Text" mode implies converting the extractor's output to plain text for comparison against the ground truth.}}
% \resizebox{\linewidth}{!}{%
\begin{tabular}{llcccc}
\toprule
Name & Mode & All & Simple & Mid & Hard \\
\midrule
Trafilatura~\citep{barbaresi2021trafilatura} & TEXT & 0.7832 & 0.8120 & 0.7782 & 0.7610 \\
Trafilatura & Html+TEXT & 0.7790 & 0.7895 & 0.7756 & 0.7732 \\
Readability~\citep{MozillaReadabilityJS} & Html+TEXT & 0.7643 & 0.7750 & 0.7598 & 0.7594 \\
magic-html~\citep{magichtml} & Html+TEXT & 0.7509 & 0.7788 & 0.7575 & 0.7144 \\
Resiliparse~\citep{bevendorff:2018} & TEXT & 0.7225 & 0.7697 & 0.7052 & 0.6985 \\
Goose3~\citep{goose3} & TEXT & 0.7047 & 0.7142 & 0.7130 & 0.6841 \\
news-please~\citep{Hamborg2017} & TEXT & 0.7046 & 0.7067 & 0.7094 & 0.6961 \\
jusText~\citep{pomikalek2011removing} & TEXT & 0.6934 & 0.7446 & 0.6965 & 0.6382 \\
BoilerPy3~\citep{boilerpy3} & TEXT & 0.6221 & 0.6481 & 0.6468 & 0.5631 \\
html2text & Html+TEXT & 0.6128 & 0.7265 & 0.6154 & 0.4958 \\
BoilerPy3 & Html+TEXT & 0.6009 & 0.6531 & 0.6030 & 0.5460 \\
GNE~\citep{gne} & Html+TEXT & 0.5181 & 0.5159 & 0.5079 & 0.5339 \\
ReaderLM-v2~\citep{wang2025readerlm} & TEXT & 0.3078 & 0.3737 & 0.2921 & 0.2630 \\
\midrule
\rowcolor{lightgreen}
Dripper & Html+TEXT & \textbf{0.8156} & \textbf{0.8382} & \textbf{0.8213} & \textbf{0.7856} \\
\rowcolor{lightgreen}
Dripper\_fallback & Html+TEXT & \textbf{0.8317} & \textbf{0.8505} & \textbf{0.8338} & \textbf{0.8101} \\
\bottomrule
\end{tabular}
% }
\label{tab:wceb_results}
\end{table}

\subsection{Ablation Study}
\label{sec:ablation}

To validate the effectiveness of our core design choices, we conducted a comprehensive ablation study focusing on the impact of training data scale, the necessity of constrained decoding, and the efficiency of our custom output representation. All models in this section utilize the Qwen3-0.6B architecture and were trained for 4 epochs.

\begin{figure}[h]
  \centering

  \includegraphics[width=0.7\linewidth]{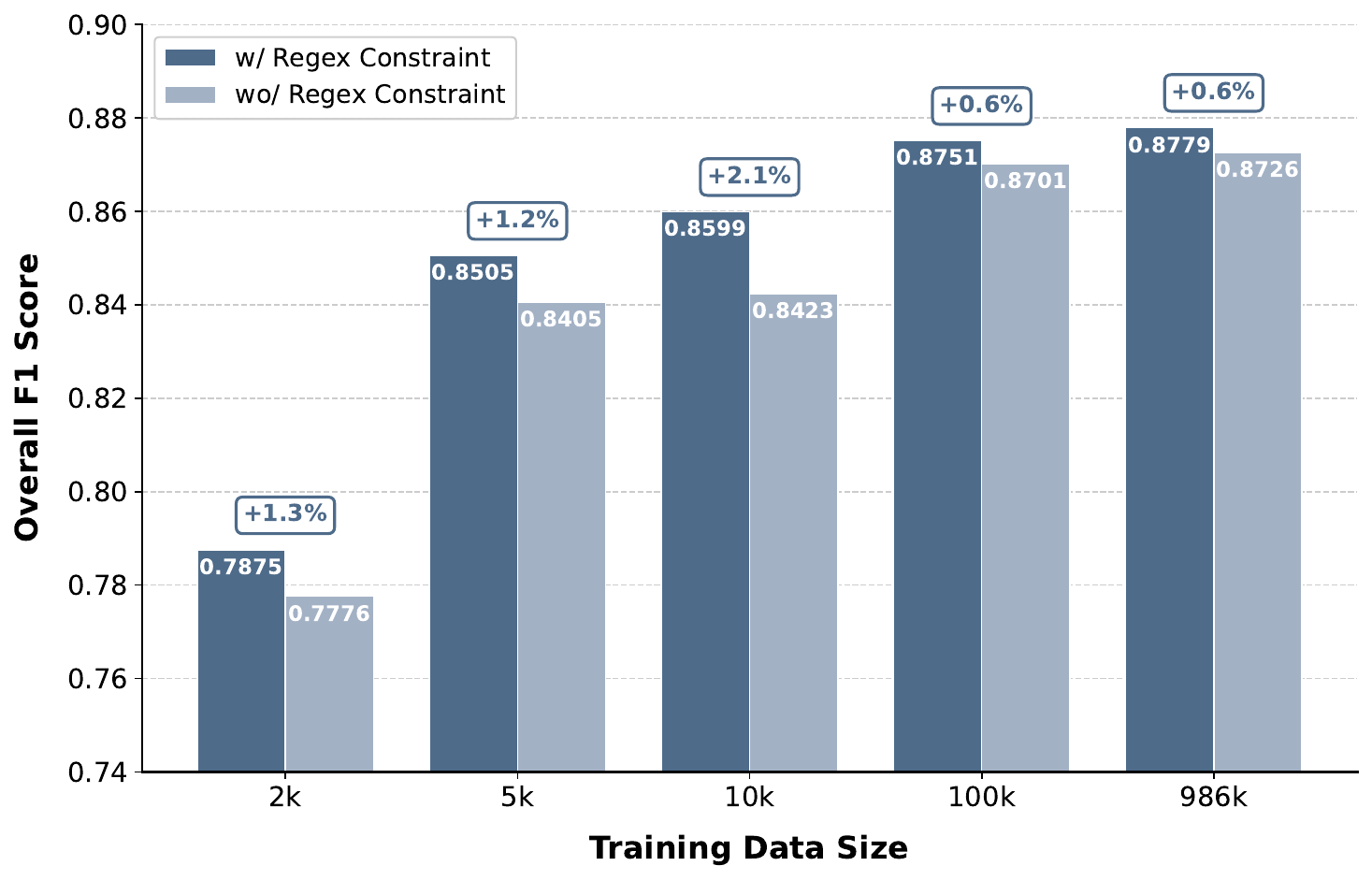}

  \caption{The figure presents the overall F1 scores for models trained with (w/ Regex Constraint) and without (wo/ Regex Constraint) the regular expression constraint across varying training data sizes (from 2k to 986k). The relative performance improvement percentages are indicated by the badges above each pair of bars. The constraint consistently enhances model performance, particularly in low-data regimes.}
 
  \label{fig:ablation-study}
\end{figure}

We first investigate the scaling behavior of the model by varying the training set size from 2k to the full 986k samples, while simultaneously evaluating the contribution of the Regex-based constrained decoding. As presented in Figure~\ref{fig:ablation-study}, the model's performance exhibits a clear positive correlation with data scale, with the overall F1 score consistently rising from 0.7875 at 2k samples to 0.8779 at the full 986k scale. Across all data regimes, the constrained decoding mechanism consistently guarantees higher performance and structural stability, ensuring that the output remains strictly compliant with the schema and free from structural hallucinations, regardless of the data scale.

Furthermore, we evaluate the efficiency of our custom "Compact" output format (e.g., 1main2other) against the Standard JSON representation (e.g., \{"1": "main", "2": "other"\}). As shown in Table~\ref{tab:ablation_format}, the choice of output format significantly impacts system efficiency. By adopting the Compact format, we reduce the total inference time from 3301 seconds to 2530 seconds, achieving a substantial \textbf{23.3\%}reduction in latency. While applying regex constraints on top of this format incurs a marginal time cost, it boosts the F1 score from 0.8726 to 0.8779. This slight latency overhead is a negligible price to pay for the significant gains in precision and structural stability.

\begin{table}[h!]
\centering
\small
\caption{\textbf{Efficiency Analysis of Output Representations.} Comparison of inference time and accuracy. The Compact format significantly reduces latency compared to JSON, while the Regex constraint improves accuracy.}
\setlength{\tabcolsep}{6pt}
\begin{tabular}{ll|c|c}
\toprule
Output Format & Constraint & Time (s) $\downarrow$ & Overall F1 $\uparrow$ \\
\midrule
Standard JSON & wo/ Regex & 3301 & 0.8772 \\
Compact (Ours) & wo/ Regex & 2158 & 0.8726 \\
\rowcolor{lightgreen}
\textbf{Compact (Ours)} & \textbf{w/ Regex} & \textbf{2530} & \textbf{0.8779} \\
\bottomrule
\end{tabular}
\label{tab:ablation_format}
\end{table}

\subsection{Impact on Downstream Pre-training}
\label{sec:downstream_pretraining}

To empirically demonstrate the foundational value of our tool in LLM development, we conduct a controlled pre-training experiment. Our core hypothesis posits that superior HTML extraction captures richer semantic context while minimizing noise, thereby directly enhancing the sample efficiency and final performance of foundation models.

We construct two parallel pre-training corpora sourced from identical snapshots of Common Crawl. To rigorously isolate the impact of the extraction method, the first corpus is processed using Trafilatura, representing the standard heuristic approach, while the second is processed using our Dripper framework.

To ensure strict comparability, both raw collections are subjected to an identical, multi-stage post-processing regimen aligned with established practices from RefinedWeb. This pipeline enforces rigorous quality standards through five specific steps: (1) exact deduplication via SHA-256 hashing; (2) language identification and filtering using FastText classifiers; (3) heuristic quality filtering adopting the Gopher rule set; (4) safety sanitization via comprehensive URL and keyword blocklists; and (5) fuzzy deduplication leveraging MinHash with Locality-Sensitive Hashing (LSH). This standardized sanitation process results in two strictly comparable datasets of approximately 63 billion tokens each, ensuring that any downstream performance divergence can be attributed solely to the extraction quality rather than filtering discrepancies.

We then train two 1B-parameter decoder-only Transformer models from scratch on these respective corpora under identical hyperparameters. For broader benchmarking context, we also compare our results against models trained on widely-adopted open-source corpora, namely RefinedWeb~\citep{refinedweb} and FineWeb~\citep{fineweb}.

To comprehensively evaluate the impact of extraction quality on downstream performance, we assessed the models on 13 benchmarks across 3 categories using the \texttt{lm-evaluation-harness} framework. These tasks were selected to rigorously test capabilities that benefit from high-fidelity structural preservation: (1) \textbf{General Knowledge}: Including MMLU\citep{mmlu}, ARC\citep{arc}, BoolQ\citep{boolq}, CommonsenseQA\citep{commonsenseqa}, and SciQ\citep{sciq}, evaluating the model's world knowledge acquisition. (2) \textbf{Reasoning}: Including HellaSwag\citep{hellaswag}, WinoGrande\citep{winogrande}, PIQA\citep{bisk2020piqa}, SIQA\citep{siqa}, and COPA\citep{gordon2012semeval}. These tasks assess logical consistency, serving as a proxy for how well the model internalized structured data. (3) \textbf{Reading Comprehension}: Including CoQA\citep{reddy2019coqa}, OpenBookQA\citep{openbookqa}, and LAMBADA\citep{paperno2016lambada}, which measure the model's ability to process long-range dependencies and document structure.

\begin{figure}[h]
    \centering
    \includegraphics[width=0.7\linewidth]{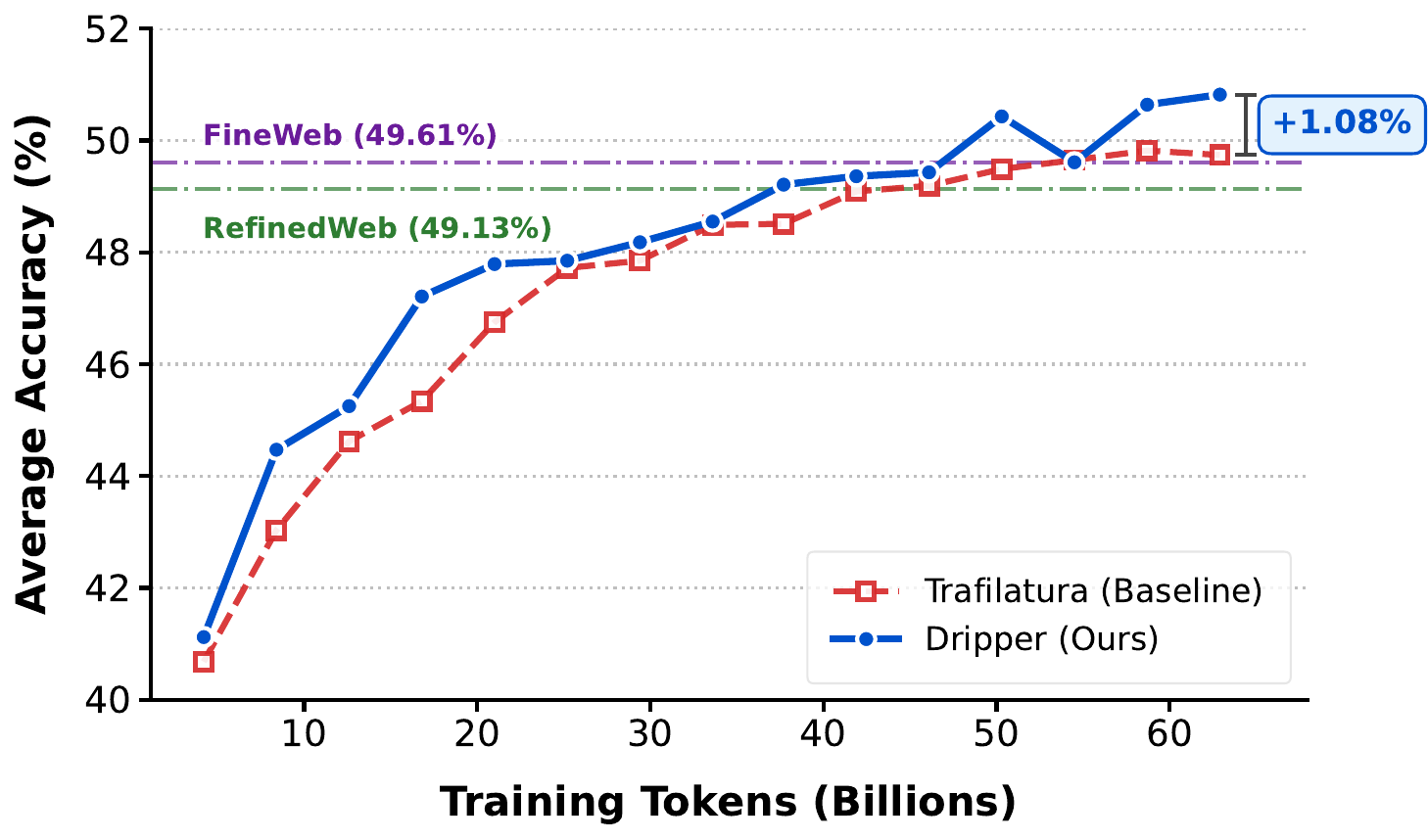} 
    
    \caption{\textbf{Impact of extraction method on pre-training dynamics.} 
    We compare the average zero-shot accuracy across 13 standard benchmarks throughout the training of 1B-parameter models on 63B tokens. 
    The model trained on \textbf{Dripper}-extracted data (blue) consistently outperforms the \textbf{Trafilatura} baseline (red) and surpasses the final performance of widely-adopted corpora RefinedWeb and FineWeb (+1.08\% improvement over baseline).}
    \label{fig:training_dynamics}
\end{figure}

\begin{table}[h!]
\centering
\caption{\textbf{Downstream evaluation results (Avg. Acc. \%) at 63B tokens.} Comparisons against Trafilatura and SOTA datasets. Dripper achieves the highest average performance.}
\label{tab:downstream_results}

\footnotesize 
\setlength{\tabcolsep}{6pt}

\begin{tabular}{l|c|ccc}
\toprule
\textbf{Training Corpus} & \textbf{Avg.} & \textbf{Gen. Know.} & \textbf{Reasoning} & \textbf{Reading} \\
\midrule
RefinedWeb & 49.13 & 45.32 & 59.10 & 40.55 \\
FineWeb       & 49.61 & 46.86 & \textbf{60.69} & 36.68 \\
\midrule
Trafilatura                      & 49.74 & 45.61 & 59.34 & 42.02 \\
\textbf{Ours (Dripper)}       & \textbf{50.82} & \textbf{47.54} & 59.83 & \textbf{42.37} \\
\bottomrule
\end{tabular}
\end{table}

The evaluation results, visualized in Figure~\ref{fig:training_dynamics}, reveal a consistent performance advantage for the model trained on Dripper-extracted data throughout the entire training trajectory. At the final checkpoint, the Dripper-based model achieves an average accuracy of \textbf{50.82\%} across 13 standard benchmarks, surpassing the Trafilatura-based baseline (49.74\%) by a significant margin of 1.08 percentage points. Since the source data distributions and cleaning procedures were controlled, this performance uplift is directly attributable to the superior quality of the structural extraction. By effectively preserving valid semantic segments that heuristics often discard, Dripper provides a richer and more coherent training signal.

\begin{figure}[t]
    \centering
    \includegraphics[width=0.7\linewidth]{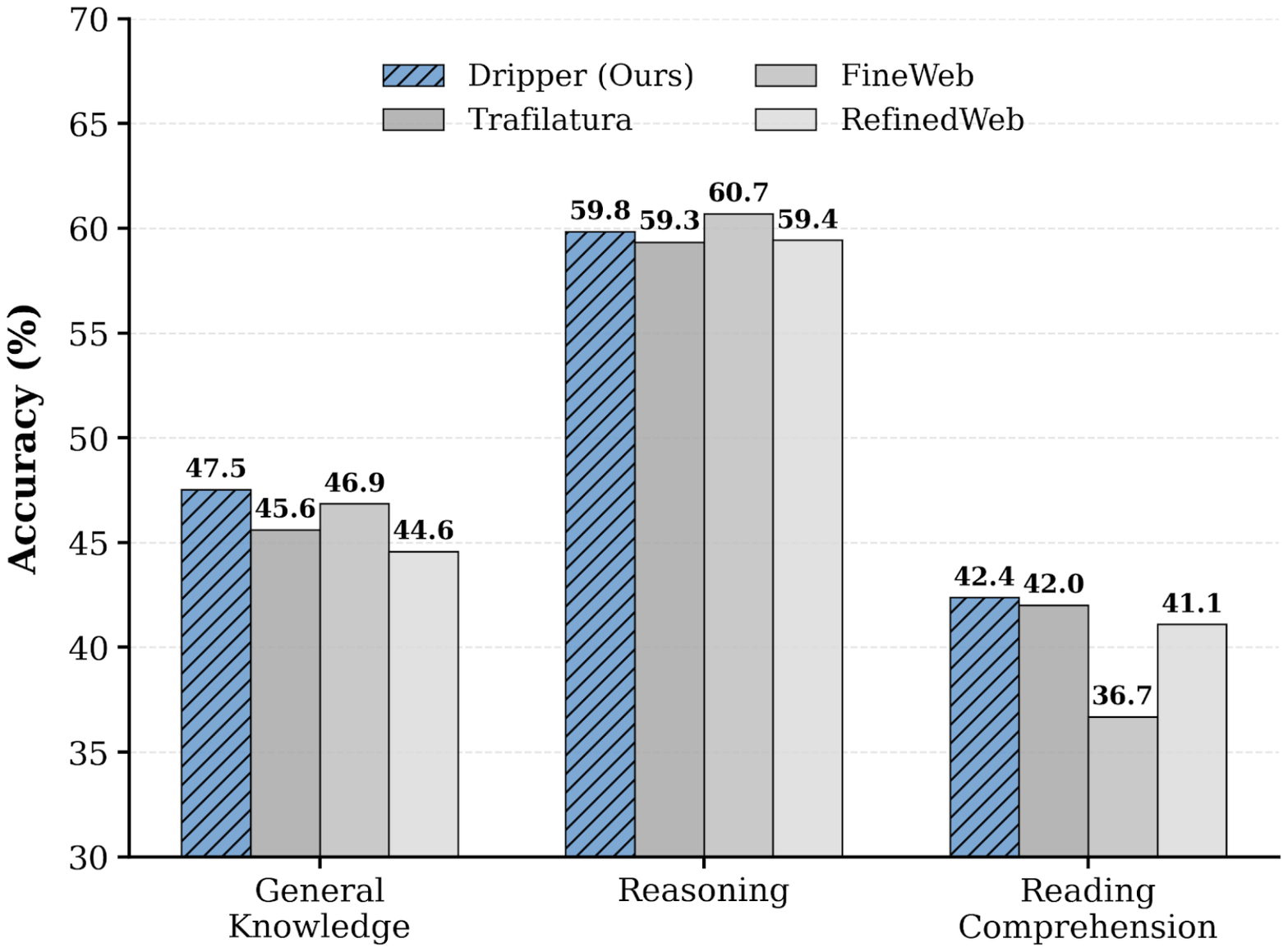}
    
    \caption{\textbf{Performance breakdown by task category.} 
    We compare the zero-shot accuracy of the \textbf{Dripper}-trained model against baselines across three core domains. 
    Dripper (blue) achieves the highest performance in \textit{General Knowledge} and \textit{Reading Comprehension}, significantly outperforming the Trafilatura baseline and other corpora (muted colors), demonstrating the efficacy of structure-aware extraction.}
    \label{fig:task_breakdown}
\end{figure}

Furthermore, a granular analysis across task categories, as illustrated in Figure~\ref{fig:task_breakdown}, highlights specific strengths of our approach. The Dripper-based model demonstrates exceptional performance in \textbf{General Knowledge tasks}, achieving 47.54\% accuracy compared to 45.61\% for the baseline. It also shows distinct gains in Reading Comprehension and Reasoning tasks. Remarkably, the model trained on our data outperforms those trained on established high-quality corpora-including RefinedWeb (49.13\%) and FineWeb (49.61\%)-in terms of overall average accuracy. These findings validate that high-fidelity content extraction serves as a critical infrastructure for next-generation corpus construction, capable of rivaling or exceeding the benefits of aggressive filtering strategies.

\section{Conclusion}
\label{sec:conclusion}

We present Dripper, a framework that bridges the gap between heuristic speed and LLM semantic reasoning via constrained sequence labeling. We demonstrate that Dripper-0.6B matches the fidelity of frontier models while mitigating hallucination risks and deployment costs. Beyond the model, we introduce WebMainBench to standardize evaluation and empirically prove that superior extraction directly enhances downstream pre-training performance. By open-sourcing our code, we aim to provide a robust infrastructure for high-quality data processing in the LLM era.

\section{Acknowledgments}
This work was supported by Shanghai Artificial Intelligence Laboratory.

\bibliographystyle{plain}
\bibliography{paper}

\clearpage
\appendix
\section{Computational Cost Estimation Details}
\label{app:cost_estimation}

For a typical decoder-only Transformer model with $L$ attention layers and hidden dimension $d$, we estimate the computational cost separately for the prefill phase (processing the input prompt) and the decode phase (autoregressive token generation). The cost is measured in floating-point operations (FLOPs). We focus on the dominant operations-matrix multiplications in linear projections and attention mechanisms-while omitting comparatively minor contributions from softmax, layer normalization, embeddings, and the language modeling head (typically accounting for less than 5\% of total FLOPs).

Let $N$ denote the length of the input prompt and $M$ the number of output tokens to be generated. The total number of model parameters is approximated as $P \approx 12 L d^2$, reflecting $4d^2$ parameters per layer for the self-attention projections ($W_Q, W_K, W_V, W_O$) and $8d^2$ for the feed-forward network (assuming a standard FFN size of $4d$).

\subsection{Prefill Phase}
In the prefill phase, the entire input sequence of length $N$ is processed in a parallel forward pass:
\begin{itemize}
    \item \textbf{Linear transformations} (attention projections and feed-forward layers): approximately $24 d^2$ FLOPs per layer per token, yielding a total of $24 L d^2 N$.
    \item \textbf{Self-attention}: each layer performs $QK^T$ and attention-value multiplications, contributing $2 d N^2$ FLOPs each, for a total of $4 L d N^2$.
\end{itemize}
Thus, the prefill phase requires:
\begin{equation}
\text{Cost}_{\text{prefill}} \approx 24 L d^2 N + 4 L d N^2
\end{equation}

\subsection{Decode Phase}
During autoregressive decoding, $M$ tokens are generated sequentially, leveraging the KV cache to avoid recomputing past keys and values:
\begin{itemize}
    \item \textbf{Linear transformations}: each generated token undergoes a full forward pass, incurring $24 L d^2$ FLOPs per token, for a total of $24 L d^2 M$.
    \item \textbf{Self-attention}: for the $i$-th generated token (where $i$ ranges from $0$ to $M-1$), the attention context length is $N + i$. Each layer therefore performs $4 d (N + i)$ FLOPs. Summing over all $M$ tokens gives:
    \[
    \sum_{i=0}^{M-1} 4 L d (N + i) = 4 L d \left( M N + \sum_{i=0}^{M-1} i \right) = 4 L d \left( M N + \frac{M(M-1)}{2} \right)
    \]
    For large $M$, this is approximated as $4 L d \left( M N + \frac{M^2}{2} \right)$.
\end{itemize}
The decode phase therefore requires approximately:
\begin{equation}
\text{Cost}_{\text{decode}} \approx 24 L d^2 M + 4 L d \left( MN + \frac{M^2}{2} \right)
\end{equation}

\subsection{Total Computational Cost}
Summing the costs of both phases yields the total estimated FLOPs used in Equation~\ref{equation}:
\begin{equation}
\begin{aligned}
\text{Total FLOPs} &= \text{Cost}_{\text{prefill}} + \text{Cost}_{\text{decode}} \\
&\approx (24 L d^2 N + 4 L d N^2) + \left( 24 L d^2 M + 4 L d MN + 2 L d M^2 \right) \\
&= 4 L d \left( N^2 + M N + \frac{M^2}{2} \right) + 24 L d^2 (N + M)
\end{aligned}
\end{equation}
The first term captures the quadratic attention costs, while the second term reflects the linear transformations dominated by model parameters.

\section{Benchmark Construction}
\label{appdx:benchmark}
\textbf{Data Sampling.} WebMainBench is constructed using a hybrid sampling strategy to ensure both broad representation and relevance. 90\% of the samples are randomly drawn from the Common Crawl dataset to cover the long-tail web, while the remaining 10\% are sampled from a list of top-ranking websites (Chinaz Alexa\footnote{https://malexa.chinaz.com/}) to include popular, professionally designed pages. The final benchmark is highly diverse, containing pages from 5,434 unique domains and 5,904 unique second-level domains.

\textbf{Annotation Rules.} To address the ambiguity in defining "main content" for unconventional layouts, we establish two core annotation principles. First, \textbf{Contextual Integrity} dictates that content integral to the main article-such as a table of contents, abstract, or reference list-is included. Conversely, contextually independent elements like "related articles" sidebars or copyright footers are excluded. Second, the main content is defined as \textbf{Human-Generated Content}, including article bodies, user comments, and Q\&A posts, while associated auto-generated metadata like usernames and timestamps are excluded.

\textbf{Annotation Process.} The annotation for each page followed a rigorous three-stage process using a custom-built tool(see Appendix, Figure \ref{snapshot}) that allowed for tag-level granularity. The process involved: (1) an initial pass by one annotator, (2) a review and correction pass by a second annotator, and (3) a final quality assurance check by a senior inspector, who made the final adjudication to resolve any discrepancies. Pages uninterpretable due to rendering issues were discarded.

\textbf{Metadata Annotation.} To enable detailed, fine-grained analysis, we annotate each page with a rich set of metadata. This includes \textbf{Language}, identified by GPT-5\citep{2025gpt5} and labeled as en (English) or non\_en (other), and \textbf{Style}, classified by GPT-5 as Conversational for pages with user-generated content or Normal otherwise. We also develop a quantitative \textbf{Difficulty Level}, determined by an \textbf{overall\_complexity\_score} calculated for each page. To compute this score, we first measure four distinct metrics: \textbf{DOM structural complexity} (based on tree depth and width), \textbf{text distribution sparsity} (transitions between text/non-text nodes), \textbf{content-type diversity} (a count of rich content types), and \textbf{link density} (the ratio of hyperlinked text). These four values are individually normalized, and their weighted sum produces the final score. Based on the distribution of this overall\_complexity\_score across the benchmark, we then categorize pages into \textbf{simple, medium, and hard} using the 30th and 70th percentiles as dynamic thresholds. Finally, we add Rich Content Tags to identify the presence of tables (<table>), code blocks (<code>), and mathematical formulas (<math> or LaTeX patterns) using BeautifulSoup.

\clearpage
\section{Baseline Methods for Web Content Extraction}
\label{app:baselines}

To evaluate the effectiveness of our proposed framework, we compare it against a diverse set of baseline methods ranging from traditional heuristics to modern neural approaches. Table~\ref{tab:baselines} provides a detailed overview of these methods, categorized by their underlying mechanisms.

\textbf{Heuristic and Rule-Based Methods} rely on statistical features (e.g., text density, link density) or DOM-specific rules to identify main content. While computationally efficient, they often struggle with the structural heterogeneity of modern web pages.
\textbf{Supervised Learning Methods} treat extraction as a classification problem, utilizing features derived from the DOM tree.
\textbf{Hybrid Systems} (e.g., Trafilatura) combine multiple heuristics and fallback strategies to improve robustness.
Finally, recent \textbf{Pre-trained Language Models} like ReaderLM-v2 attempt to leverage semantic understanding for extraction, though often at a higher computational cost.

\begin{table*}[h!]
\centering
\small  
\caption{An overview of the baseline methods for web content extraction used in our comparative analysis.}
\begin{tabular}{@{}ll@{}}
\toprule
\textbf{Method} & \textbf{Description} \\
\midrule
\multicolumn{2}{@{}l}{\textit{\textbf{Heuristic and Rule-Based Methods}}} \\
\midrule
Readability & Reader view algorithm that removes distracting elements based on tag and class scores \\
jusText & Two-pass algorithm using block size, link density, and stopword heuristics \\
Goose3 & Article extractor relying on hand-crafted rules and stopword density \\
html2text & A simple converter that transforms HTML directly to Markdown without filtering \\
GNE & General News Extractor based on text and symbol density using mathematical formulas \\
Resiliparse & Fast and robust heuristic extractor optimized for web archiving \\
\midrule
\multicolumn{2}{@{}l}{\textit{\textbf{Supervised Learning Methods}}} \\
\midrule
BoilerPy3 & Python port of Boilerpipe, utilizing decision trees for text block classification \\
\midrule
\multicolumn{2}{@{}l}{\textit{\textbf{Hybrid Systems and Production Tools}}} \\
\midrule
Trafilatura & Sophisticated cascade of rules with jusText and Readability as robust fallbacks \\
news-please & A meta-extractor that combines results from multiple underlying extractors \\
magic-html & A tool focused on HTML structure simplification for downstream pipelines \\
\midrule
\multicolumn{2}{@{}l}{\textit{\textbf{Pre-trained Language Models}}} \\
\midrule
ReaderLM-v2  & SLM-based extractor fine-tuned to translate raw HTML to clean Markdown \\
\bottomrule
\end{tabular}
\label{tab:baselines}
\end{table*}

\section{Standard Benchmarks Details}
\label{app:wceb}

In addition to our proposed WebMainBench, we utilize the Web Content Extraction Benchmark (WCEB) to assess generalization capabilities on legacy data. WCEB is a consolidated collection of several historically significant datasets, ensuring a fair comparison with prior literature. Table~\ref{tab:datasets} details the composition of WCEB, which aggregates over 3,000 pages from diverse sources, ranging from news articles (CleanPortalEval) to blogs (Dragnet) and raw web crawls (Common Crawl). Note that unlike WebMainBench, most of these datasets provide ground truth primarily in plain text format.

\begin{table}[!htbp]
\centering
\caption{Composition of the Web Content Extraction Benchmark (WCEB), aggregating multiple legacy datasets.}
\small
\setlength{\tabcolsep}{5pt}
\begin{tabularx}{\linewidth}{@{}l r X@{}} 
\toprule
\textbf{Dataset} & \textbf{Pages} & \textbf{Source \& Characteristics} \\
\midrule
CleanEval & 738 & The de-facto standard from the 2007 shared task, combining development and evaluation sets with basic structural markup. \\
CleanPortalEval & 71 & An extension of CleanEval featuring multi-page samples from 4 major news domains. \\
CETD & 700 & A dataset created specifically for evaluating density-based extractors across 6 domains. \\
Dragnet & 1,379 & A large collection from RSS feeds, 23 major news sites, and 178 blogs. \\
L3S-GN1 & 621 & Created by BoilerPipe authors, featuring unique span-wrapped CSS class annotations for 5-level content relevance. \\
Google-Trends-2017 & 180 & Created for training the BoilerNet neural network, featuring binary CSS class annotations on DOM leaf nodes. \\
Readability & 115 & The official test suite for Mozilla's Readability.js, containing original and simplified HTML pairs. \\
Scrapinghub & 181 & A benchmark dataset created by Zyte for evaluating proprietary extraction services. \\
\bottomrule
\end{tabularx}
\label{tab:datasets}
\end{table}

\clearpage
\section{Annotation Tool Interface}
\label{app:annotation_tool}

\begin{figure*}[h]
    \centering
    \includegraphics[width=0.9\linewidth]{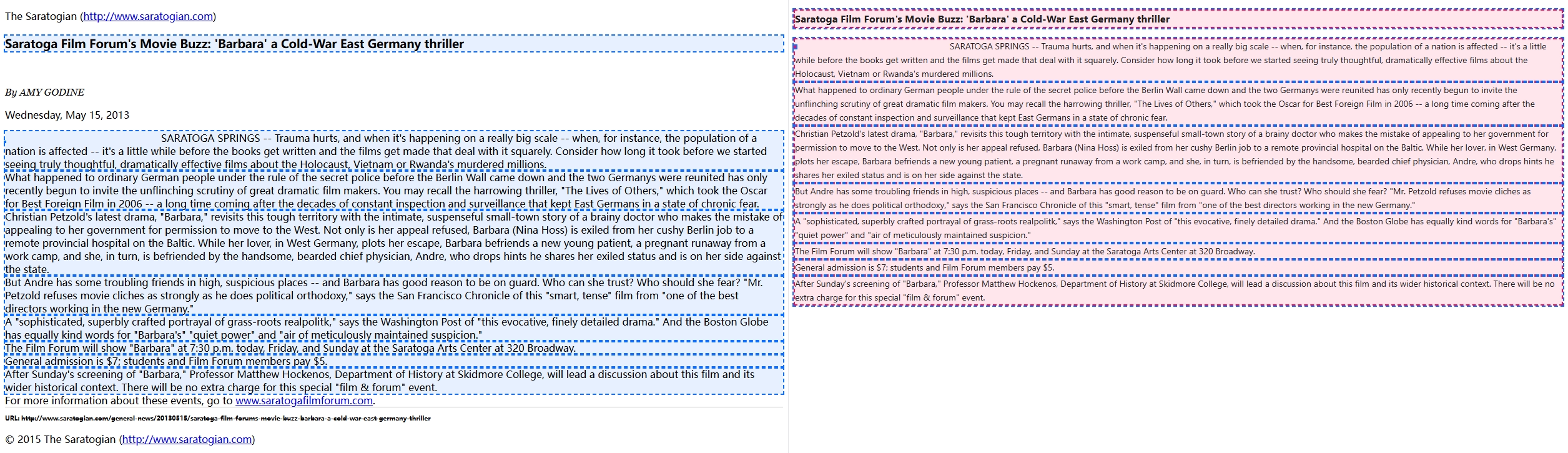}
    \caption{Screenshot of the custom web page annotation tool. The user selects main content blocks on the left (highlighted in blue), while the right pane shows the real-time extraction result.}
    \label{snapshot}
\end{figure*}

To construct the high-precision WebMainBench dataset, we developed a custom web-based annotation tool designed for efficiency and granularity. As shown in Figure~\ref{snapshot}, the interface features a dual-pane design:
\begin{itemize}
    \item \textbf{Left Pane (DOM View):} Displays the rendered web page. Annotators can hover over elements to inspect their boundaries and click to select/deselect them. Selected elements (constituting the main content) are highlighted in blue.
    \item \textbf{Right Pane (Preview):} Provides a real-time rendered preview of the extracted content based on the current selection. This allows annotators to instantly verify the integrity and readability of the extraction.
\end{itemize}
This tool operates directly on the HTML DOM tree, ensuring that the ground truth is captured as a precise set of HTML tags rather than unstructured text.

\clearpage
\section{WebMainBench Data Schema}
\label{app:benchmark_example}

WebMainBench provides rich metadata and structural ground truth to facilitate fine-grained analysis. Figure~\ref{fig:benchmark_example} illustrates a typical sample from the dataset in JSON format. Key fields include:
\begin{itemize}
    \item \texttt{html}: The raw HTML source, potentially containing `cc-select` attributes injected by the annotation tool to mark ground-truth elements.
    \item \texttt{main\_html}: The clean, ground-truth HTML containing only the selected main content subtree.
    \item \texttt{convert\_main\_content}: The canonical Markdown conversion of the main content, used for text-level evaluation.
    \item \texttt{meta}: A rich set of attributes describing the page, including its difficulty level (\textit{easy, mid, hard}), page style (\textit{Normal, Forum, etc.}), and flags for rich content types (tables, code, equations).
\end{itemize}

\begin{figure*}[!htbp]
  \centering
  \begin{lstlisting}[
    language=json,
    basicstyle=\small\ttfamily,
    frame=single,
    numbers=left,
    breaklines=true,
    showstringspaces=false,
    columns=flexible
  ]
{
  "track_id": "CC-MAIN-2023-50-...",
  "html": "<html><body><h1 cc-select=True>Hello world!</h1><aside>advertisement</aside></body></html>",
  "main_html": "<html><body><h1>Hello world!</h1></body></html>",
  "convert_main_content": "# Hello world!",
  "meta": {
    "language": "en",
    "style": "Normal",
    "level": "easy",
    "table": "without",
    "code": "without",
    "equation": "without"
  }
}
  \end{lstlisting}
  \caption{An illustrative data sample from WebMainBench. The schema includes the raw source, the structural ground-truth (main\_html), a canonical Markdown conversion for standardized evaluation, and a rich set of metadata tags to support fine-grained difficulty analysis.}
  \label{fig:benchmark_example}
\end{figure*}

\section{Detailed Benchmark Statistics}
\label{app:benchmark_stats}

In this section, we provide granular statistics regarding the composition of WebMainBench to demonstrate its diversity and coverage. WebMainBench consists of 7,809 samples. As detailed in Section 4.2.1, the composition follows a hybrid sampling strategy: 90\% are randomly sampled from Common Crawl to capture the ``long-tail'' of the web, while 10\% are sampled from top-ranking websites to ensure the inclusion of popular, high-quality pages.

\textbf{Domain Diversity.} The dataset covers 5,434 unique domains, confirming that the data is not dominated by a few sources but possesses a high degree of diversity. Table \ref{tab:domain_stats} lists the top 10 domains sorted by sample count. Furthermore, the benchmark spans 150 distinct Top-Level Domains (TLDs), indicating a broad spectrum of global regions and website categories. The distribution of the top 10 TLDs is presented in Table \ref{tab:tld_dist}.

\begin{figure*}[t] 
    \centering
    \includegraphics[width=\textwidth]{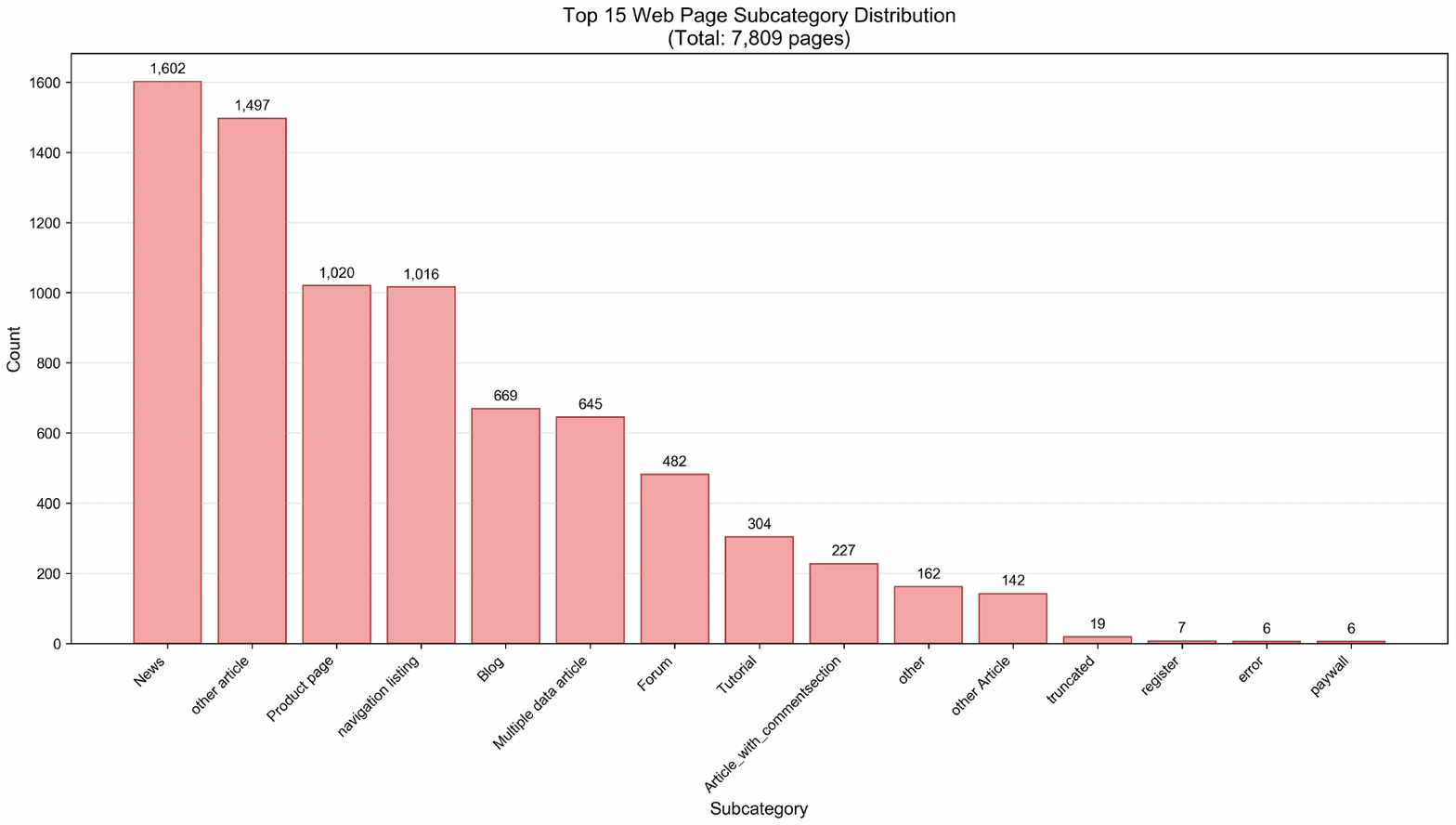}
    \caption{Top 15 Web Page Subcategory Distribution. The types were classified semantically using GPT-5.}
    \label{fig:subcategory_dist}
\end{figure*}

\textbf{Page Category Distribution.} We utilized GPT-5 to classify the semantic type of every page in the benchmark. As visually demonstrated in Figure \ref{fig:subcategory_dist}, the dataset covers a diverse range of page layouts, ranging from standard news articles to forums and product pages.

\textbf{Language Diversity.} The dataset includes web pages in 46 different languages. We present the partial language statistics (top 10) in Table \ref{tab:lang_dist}. 

\begin{table*}[h]
    \centering
  
    \caption{Partial Domain Statistics (Top 10 Sorted by Sample Count). This table highlights the variety in page styles, difficulty levels, and rich content elements even within the most frequent domains.}
    \label{tab:domain_stats}
    \resizebox{\textwidth}{!}{%
    \begin{tabular}{lcccccccc}
        \toprule
        \textbf{Domain} & \textbf{Count} & \textbf{Percent} & \textbf{Lang} & \textbf{Style} & \textbf{Level} & \textbf{Table} & \textbf{Code} & \textbf{Eq.} \\
        \midrule
        aniruddhadeb.com & 39 & 0.49\% & en & Article & simple & 1 & 9 & 36 \\
        politics.stackexchange.com & 30 & 0.38\% & en & Forum & mid & 0 & 0 & 0 \\
        www.ask.com & 29 & 0.37\% & en & Article & simple & 1 & 0 & 3 \\
        en.wikipedia.org & 27 & 0.34\% & en & Article & hard & 20 & 1 & 0 \\
        www.china.org.cn & 23 & 0.29\% & en & Article & simple & 21 & 0 & 0 \\
        money.cnn.com & 22 & 0.28\% & en & Article & hard & 18 & 0 & 7 \\
        data.epo.org & 21 & 0.27\% & en & Article & simple & 21 & 0 & 0 \\
        m.weibo.cn & 19 & 0.24\% & zh & Forum & simple & 0 & 0 & 0 \\
        spanish.china.org.cn & 15 & 0.19\% & es & Article & simple & 14 & 0 & 0 \\
        china.org.cn & 14 & 0.18\% & en & Article & mid & 13 & 0 & 0 \\
        \bottomrule
    \end{tabular}%
    }
\end{table*}

\begin{table*}[h]
    \centering
    \begin{minipage}[t]{0.48\textwidth}
        \centering
        \caption{Partial Top-Level Domain (TLD) Distribution (Top 10).}
        \label{tab:tld_dist}
        \begin{tabular}{lrr}
            \toprule
            \textbf{TLD} & \textbf{Count} & \textbf{Percent} \\
            \midrule
            com & 4550 & 57.69\% \\
            org & 816 & 10.35\% \\
            cn & 459 & 5.82\% \\
            net & 318 & 4.03\% \\
            uk & 235 & 2.98\% \\
            edu & 180 & 2.28\% \\
            de & 101 & 1.28\% \\
            au & 94 & 1.19\% \\
            ru & 69 & 0.87\% \\
            gov & 59 & 0.75\% \\
            \bottomrule
        \end{tabular}
    \end{minipage}
    \hfill

    \begin{minipage}[t]{0.48\textwidth}
        \centering
        \caption{Partial Language Distribution (Top 10) in the benchmark.}
        \label{tab:lang_dist}
        \begin{tabular}{lrr}
            \toprule
            \textbf{Language} & \textbf{Count} & \textbf{Percent} \\
            \midrule
            English & 6711 & 85.09\% \\
            Chinese & 716 & 9.08\% \\
            Spanish & 61 & 0.77\% \\
            German & 51 & 0.65\% \\
            Japanese & 48 & 0.61\% \\
            Russian & 45 & 0.57\% \\
            French & 36 & 0.46\% \\
            Italian & 22 & 0.28\% \\
            Korean & 20 & 0.25\% \\
            Portuguese & 17 & 0.22\% \\
            \bottomrule
        \end{tabular}
    \end{minipage}
\end{table*}

\begin{figure*}[t!]
    \centering
    \begin{lstlisting}[
  language=Python,
  basicstyle=\scriptsize\ttfamily,     % 改成 scriptsize（比footnotesize 再小一号）
  frame=single,
  framesep=3pt,                        % 框内边距更小
  framerule=0.4pt,
  breaklines=true,
  breakatwhitespace=true,
  numbers=none,
  aboveskip=1pt,
  belowskip=1pt,
  xleftmargin=2pt,
  xrightmargin=2pt,
  framexleftmargin=2pt,
  framexrightmargin=2pt,
  linewidth=\textwidth,
  lineskip=-1pt,                       % ← 关键：强制缩小行间距
]
f"""[Role and Task Objectives]
You are the most precise web content extraction engine. Your goal is to act as a ruthless noise filter, removing all auxiliary content and UI elements not directly related to the core topic of the page to achieve the highest recall and precision. Professional Constraint: Your judgment must be based on HTML structure and content relevance, not visual presentation.
You will see a simplified HTML structure where each node has a unique _item_id attribute. Your task is to determine whether each node belongs to the "main" content or "other" content, preserving the body of the webpage and removing irrelevant parts like navigation bars and metadata.

[Output Requirements]
Only return results in JSON format, as shown in the example:
{{ "1": "main", "2": "other", "3": "main" }}
[!] Do not include any explanation in the output-return only JSON data.
[!] If unsure whether a node belongs to the main content, mark it as "other" by default.
[!] You must accurately distinguish between main and other content.

[Emotional Encouragement]
Let's conquer this content classification challenge together! You have sharp insight and can precisely identify the core value of a webpage. Remember, your judgment helps users quickly access the most valuable information-this is highly meaningful work!

[Correct Procedure for Main Content Extraction]
Step 1: Grasp the core theme of the page.
Quickly understand the central topic or information focus; this will guide your classification judgment.
Step 2: Classify each node.
Judgment Principle: Is this node related to the core theme?
- Related -> "main"
- Unrelated -> "other"

1. Criteria for Main Content ("main"):
    a. Articles/News/Product Details/Blogs/Match Info:
    Manually edited content, including body paragraphs, titles, descriptions, embedded images, tables, code blocks, or multiple articles. Includes structured elements like abstracts, introductions, tables of contents, and citations.
    [Exclude]: Auxiliary info (e.g., view counts, likes, comment counts, social sharing buttons).

    b. Forum Discussions:
    Original post content and replies (including quoted content).
    [Include]: All user-generated text (paragraphs, quotes, images).
    [Exclude]: Usernames, timestamps, floor numbers, likes, etc.

    c. Other Content-Dense Pages:
    Any meaningful readable content (including privacy policies, legal terms, site descriptions, etc.).
    [Note]: If non-core content appears (e.g., service link at the bottom of a news page), mark as "other".
    [Exclude]: Pure link modules without metadata, or product lists on e-commerce sites without attributes.
    [Remember]: Your work filters noise to reach the core! This is a heavy but rewarding responsibility!

2. Criteria for Auxiliary Content ("other"):
    a. Navigation and UI Elements:
    Menus, sidebars, footers, breadcrumbs, pagination links, "skip to content" prompts, etc.

    b. Auxiliary Information:
    Any info describing the state of the content carrier or interaction should be "other".
    - Includes: Publish time, author name, views, likes, comment counts, floor numbers, votes, usernames, etc.
    - Exception (mark as "main"): Key attributes that directly define or describe the core object (article, product, book, service, job).
        Example: Author/ISBN for books, specs/price for products, title/salary for recruitment, time/location for event invites.
        Principle: If removing it makes the core theme incomplete, mark as "main".

    c. Ads and Promotion:
    Ads or sponsored content interspersed in articles, social sharing buttons, follow prompts, external recommendations.

    d. Recommended Content:
    Related content, trending topics, "read more", "you may also like", "next article", etc.

[Detailed Processing Guide (Summary)]
    1. Golden Rule: If the page has a clear theme, mark it "main". Annotate multiple articles individually.
    2. Special Docs: Event invites, job postings, and legal terms are "main". TOCs, abstracts, and quotes are part of the structure.
    3. Auxiliary Info: Keep ("main") manual contact info, specs, or book metadata. Remove ("other") templated carrier status info (views/time).
    4. Interaction: All "Download", "Read More", or "Click" buttons/links are "other".

Now, show your professional skills! Analyze the following HTML and return results for each _item_id:
Input HTML:
{html_str}
"""
    \end{lstlisting}
    \caption{Full prompt used for data synthesis. The instructions emphasize strict separation between core content and UI/metadata elements.}
    \label{code:prompt}
\end{figure*}

\clearpage
\section{Analysis of Classification Metrics}
\label{app:classification_metrics}

In this section, we provide a detailed analysis of the model's internal classification performance relative to the final extraction quality. In fact, we closely monitored these metrics (Precision, Recall, and F1) throughout our model development process to assess internal classification performance. We provide these results in Table \ref{tab:block_metrics}.

\begin{table}[h]
    %\centering
    \caption{Block-level classification metrics across different training data sizes.}
    \label{tab:block_metrics}
    %\resizebox{0.85\textwidth}{!}{%
    \begin{tabular}{lccc}
        \toprule
        \textbf{Data Size} & \textbf{Precision} & \textbf{Recall} & \textbf{F1} \\
        \midrule
        2k   & 0.7729 & 0.8234 & 0.7279 \\
        5k   & 0.7934 & 0.8880 & 0.7970 \\
        10k  & 0.7996 & 0.8928 & 0.8061 \\
        100k & 0.8155 & 0.9077 & 0.8278 \\
        986k & 0.8190 & 0.9109 & 0.8331 \\
        \bottomrule
    \end{tabular}%
    %}
\end{table}

While the data confirms our model's strong classification capability, we deliberately chose ROUGE-N as our primary reporting metric for three key reasons.

First, our pre-processing creates blocks with significant content length variance. A block can range from a single boilerplate word to a 2,000-word main article. Standard classification metrics treat all blocks equally, meaning a model could achieve a high F1 score by correctly classifying hundreds of tiny boilerplate blocks while missing the single, massive main content block. This would yield a high classification score but a completely failed extraction. 

Secondly, ROUGE-N better aligns with the end-user's objective, which is to obtain the complete main text. By measuring the overlap between the extracted text and the ground truth, ROUGE implicitly weights blocks by their information content, ensuring that the metric reflects the actual utility of the output.

Finally, prioritizing ROUGE-N ensures consistency with established benchmarks in the web extraction literature, where ROUGE-L or ROUGE-N are the standard metrics for comparison.

\section{Detailed Pre-processing Algorithm}
\label{app:preprocessing_details}

In this section, we provide a comprehensive description of the HTML simplification algorithm, which serves as the cornerstone of the Dripper framework. The primary goal of this algorithm is to \textbf{drastically reduce the HTML token count while preserving the critical semantic and structural cues necessary for accurate content classification}. This is achieved through a multi-stage process applied to the raw HTML.

\textbf{1. DOM Cleaning and Pruning.} We first parse the HTML and proactively remove entire subtrees known to be boilerplate. This includes tags such as \texttt{<script>}, \texttt{<style>}, \texttt{<header>}, \texttt{<footer>}, and \texttt{<nav>}. Furthermore, we heuristically remove elements whose \texttt{class} or \texttt{id} attributes contain keywords like `nav', `footer', or `header', or which have CSS styles indicating they are hidden (e.g., \texttt{display: none}).

\textbf{2. Attribute Simplification.} To reduce noise and token overhead, we strip nearly all attributes from all elements. The only exceptions are the \texttt{class} and \texttt{id} attributes, which are often the most informative semantic markers in modern web design, and for \texttt{<img>} tags, we also preserve the \texttt{src} (excluding large base64 data) and \texttt{alt} attributes.

\textbf{3. Semantic Block Segmentation.} The core of our method involves converting the cleaned DOM tree into a linear sequence of semantic blocks. We perform a recursive traversal of the DOM, segmenting it at natural block-level boundaries. Our algorithm intelligently handles mixed content:
\begin{enumerate}
    \item It identifies and preserves atomic block-level elements (e.g., a standalone paragraph or \texttt{<div>}).
    \item It aggregates consecutive inline elements (e.g., \texttt{<span>}, links with text) and unwrapped text nodes into coherent blocks, wrapping them in a custom tag if necessary to maintain structure.
    \item It makes special provisions for complex structures like tables and lists, ensuring they are treated as single, indivisible units where appropriate.
\end{enumerate}

\textbf{Parallel Generation Strategy.} A critical innovation in our pipeline is the parallel generation of Simplified HTML and Mapping HTML.Both representations undergo identical block segmentation, ensuring a one-to-one correspondence between blocks. However, Simplified HTML used for model input undergoes the full pruning and truncation process (steps 1-4). In contrast, Mapping HTML, used for final output reconstruction, undergoes only the initial cleaning (step 1) and segmentation (step 3), preserving the original, un-truncated 

Finally, we inject a unique \_item\_id attribute to each block in both the Simplified and Mapping HTML. This allows the classification labels produced by Dripper-0.6B on the simplified sequence to be precisely mapped back to the rich, original content blocks for the final extraction.

\section{Training configurations}
\label{app:training_config}
The specific hyperparameters and training configurations for supervised fine-tuning model are shown in table \ref{tab:hyperpara}.

\begin{table}[h]
    \centering
               
    \caption{Supervised Fine-tuning (SFT) Configuration Details}
    \label{tab:hyperpara}
    
    % \resizebox{\linewidth}{!}{    
    \begin{tabular}{lll}
    \toprule
    \textbf{Category} & \textbf{Parameter} & \textbf{Value} \\
    \midrule
    
    \multirow{3}{*}{Model \& Framework} & Base Model & Qwen3-0.6B \\
     & Fine-tuning Method & Full-parameter SFT \\
     & Training Framework & LLaMA-Factory\cite{zheng2024llamafactory} \\
    \midrule
    \multirow{3}{*}{Training Dynamics} & Epochs & 4 \\
     & Global Batch Size & 128 \\
     & Max Sequence Length & 32,000 \\
    \midrule
    \multirow{3}{*}{Optimizer \& Scheduler} & Learning Rate Scheduler & Cosine \\
     & Peak Learning Rate & 1.00E-04 \\
     & Warmup Ratio & 0.1 \\
    \midrule
    \multirow{2}{*}{Hardware \& Efficiency} & Hardware & 32$\times$ NVIDIA A100 GPU \\
     & Precision & BF16 \\
    \bottomrule
    \end{tabular}
\end{table}

\clearpage
\section{Domain-Specific Evaluation Results}
\label{app:domain_eval}

To comprehensively assess the generalization capabilities of our framework, we utilized the Topic and Format classifiers proposed by \cite{organize_the_web_citation_key} to categorize all 7,809 samples in the WebMainBench. Based on this classification, we calculated the ROUGE-N F1 scores for all methods across 24 distinct topics (e.g., Science \& Tech, Finance, Health) and 24 distinct formats (e.g., News Article, Tutorial, Forum).

As shown in Table \ref{tab:domain_summary}, Dripper\_fallback consistently secures the 1st place ranking \textbf{across various topic and format categories}. On average, it outperforms the strongest existing baseline (magic-html) by approximately \textbf{25\%} (up from 0.71 to 0.89), demonstrating the exceptional robustness of this strategy. The standalone Dripper model consistently ranks \textbf{second}, maintaining a significant lead over heuristic methods.

These results consistently demonstrate our model's strong capabilities across diverse domains. The complete breakdown of performance scores for every individual Topic and Format category is presented in Tables \ref{tab:formats_full} - \ref{tab:topics_full}.

\begin{table*}[h]
    \centering
    \caption{Detailed ROUGE-N F1 Scores across Formats.}
    \label{tab:formats_full}
    \setlength{\tabcolsep}{1.5pt}
    \resizebox{\textwidth}{!}{
    \begin{tabular}{lcccccccccccccccccccccccc}
        \toprule
        \textbf{Method} & \textbf{Acad.} & \textbf{About(O)} & \textbf{Cmnt} & \textbf{Struct} & \textbf{News} & \textbf{List} & \textbf{About(P)} & \textbf{Prod} & \textbf{Lstcl} & \textbf{Blog} & \textbf{Knowl} & \textbf{Creat} & \textbf{Tutor} & \textbf{Spam} & \textbf{Nonfic} & \textbf{Trscrpt} & \textbf{Doc} & \textbf{Forum} & \textbf{Review} & \textbf{Trunc} & \textbf{News(O)} & \textbf{Legal} & \textbf{FAQ} & \textbf{Supp} \\
        \midrule
        Dripper\_fallback & \textbf{0.9280} & \textbf{0.8430} & \textbf{0.8759} & \textbf{0.8419} & \textbf{0.9625} & \textbf{0.8360} & \textbf{0.8091} & \textbf{0.8419} & \textbf{0.9724} & \textbf{0.9302} & \textbf{0.9268} & \textbf{0.9455} & \textbf{0.9467} & \textbf{0.8473} & \textbf{0.9787} & \textbf{0.9659} & \textbf{0.9229} & \textbf{0.8694} & \textbf{0.8458} & 0.7948 & \textbf{0.9265} & \textbf{0.9636} & \textbf{0.9126} & \textbf{0.9044} \\
        Dripper & 0.8914 & \textbf{0.8430} & 0.8314 & 0.8173 & 0.9570 & 0.8242 & 0.7687 & 0.8319 & 0.9603 & 0.9108 & 0.9194 & 0.8902 & 0.9421 & 0.7949 & 0.9735 & 0.9457 & 0.8708 & 0.8583 & 0.8355 & \textbf{0.7948} & 0.9113 & 0.9546 & 0.8765 & 0.8634 \\
        magic-html & 0.8606 & 0.7260 & 0.4531 & 0.6084 & 0.8864 & 0.5244 & 0.6424 & 0.6451 & 0.8673 & 0.7690 & 0.8151 & 0.8248 & 0.8113 & 0.7074 & 0.9301 & 0.8881 & 0.8082 & 0.5470 & 0.5426 & 0.5127 & 0.8323 & 0.9066 & 0.5514 & 0.7604 \\
        readability & 0.7820 & 0.6516 & 0.4508 & 0.4791 & 0.8692 & 0.4603 & 0.5421 & 0.5401 & 0.8033 & 0.7281 & 0.7897 & 0.7887 & 0.7686 & 0.7030 & 0.9179 & 0.8519 & 0.7552 & 0.6228 & 0.5206 & 0.5008 & 0.7822 & 0.8916 & 0.6627 & 0.6374 \\
        trafilatura-html & 0.7955 & 0.6807 & 0.5400 & 0.4288 & 0.8334 & 0.5068 & 0.6101 & 0.5468 & 0.7814 & 0.7138 & 0.7344 & 0.7510 & 0.7314 & 0.6649 & 0.8701 & 0.8526 & 0.6028 & 0.5921 & 0.5865 & 0.4972 & 0.7463 & 0.7977 & 0.6959 & 0.6859 \\
        resiliparse & 0.7830 & 0.6447 & 0.4825 & 0.4319 & 0.7679 & 0.5090 & 0.5875 & 0.5560 & 0.7737 & 0.7453 & 0.7321 & 0.7197 & 0.7287 & 0.5712 & 0.8894 & 0.8369 & 0.6662 & 0.7163 & 0.6068 & 0.4430 & 0.6971 & 0.8034 & 0.7701 & 0.6932 \\
        trafilatura-md & 0.7807 & 0.6570 & 0.5389 & 0.4221 & 0.8272 & 0.4793 & 0.5918 & 0.5268 & 0.7772 & 0.7188 & 0.7201 & 0.7245 & 0.7135 & 0.6521 & 0.8634 & 0.8322 & 0.5510 & 0.5816 & 0.5884 & 0.4906 & 0.7438 & 0.7775 & 0.7175 & 0.6590 \\
        trafilatura-text & 0.7654 & 0.6459 & 0.5350 & 0.4001 & 0.7977 & 0.4667 & 0.5737 & 0.5083 & 0.7521 & 0.7049 & 0.6987 & 0.7041 & 0.6907 & 0.6299 & 0.8484 & 0.8214 & 0.5347 & 0.5608 & 0.5815 & 0.4794 & 0.7177 & 0.7583 & 0.6847 & 0.6308 \\
        html2text-md & 0.7767 & 0.5150 & 0.6013 & 0.6441 & 0.5860 & 0.6216 & 0.5211 & 0.4645 & 0.6909 & 0.6070 & 0.6320 & 0.7196 & 0.7000 & 0.5138 & 0.7578 & 0.8108 & 0.8293 & 0.4195 & 0.4971 & 0.3344 & 0.5511 & 0.8701 & 0.6914 & 0.6372 \\
        boilerpy3-text & 0.6770 & 0.6564 & 0.4301 & 0.2518 & 0.7798 & 0.2752 & 0.4967 & 0.5118 & 0.6640 & 0.6381 & 0.6269 & 0.6578 & 0.6272 & 0.5390 & 0.8390 & 0.7502 & 0.4333 & 0.4872 & 0.5286 & 0.4665 & 0.6895 & 0.7463 & 0.6150 & 0.6150 \\
        gne & 0.6632 & 0.6204 & 0.2916 & 0.2830 & 0.7518 & 0.2952 & 0.4540 & 0.4461 & 0.6907 & 0.5959 & 0.6246 & 0.6860 & 0.5845 & 0.5723 & 0.8065 & 0.7052 & 0.4829 & 0.4033 & 0.4318 & 0.4478 & 0.6722 & 0.7880 & 0.4510 & 0.5080 \\
        newsplease & 0.6309 & 0.5871 & 0.3626 & 0.2409 & 0.6320 & 0.3383 & 0.4593 & 0.4437 & 0.7317 & 0.6930 & 0.6392 & 0.5131 & 0.6625 & 0.3158 & 0.8476 & 0.7799 & 0.4372 & 0.6525 & 0.5328 & 0.3582 & 0.5961 & 0.7123 & 0.6744 & 0.6217 \\
        boilerpy3-html & 0.6475 & 0.4920 & 0.3969 & 0.2694 & 0.6709 & 0.2578 & 0.4240 & 0.3810 & 0.6212 & 0.5563 & 0.5347 & 0.6459 & 0.5852 & 0.4819 & 0.7643 & 0.7428 & 0.5231 & 0.3654 & 0.4172 & 0.3420 & 0.5492 & 0.8026 & 0.6178 & 0.5384 \\
        justtext & 0.6357 & 0.6536 & 0.5135 & 0.1471 & 0.5649 & 0.2704 & 0.4695 & 0.4605 & 0.7281 & 0.7168 & 0.5377 & 0.4444 & 0.6538 & 0.1885 & 0.8461 & 0.7810 & 0.3567 & 0.5653 & 0.6050 & 0.3415 & 0.5116 & 0.6823 & 0.7084 & 0.6121 \\
        goose3 & 0.5351 & 0.5816 & 0.2858 & 0.1517 & 0.6158 & 0.2247 & 0.3910 & 0.4129 & 0.6982 & 0.6181 & 0.5545 & 0.3598 & 0.6218 & 0.1686 & 0.8174 & 0.7118 & 0.3192 & 0.3404 & 0.4693 & 0.3874 & 0.5386 & 0.5077 & 0.6699 & 0.4959 \\
        readerlm & 0.3318 & 0.2753 & 0.1519 & 0.1172 & 0.3553 & 0.1306 & 0.1565 & 0.1564 & 0.3107 & 0.2321 & 0.2651 & 0.2852 & 0.2959 & 0.2356 & 0.4072 & 0.3647 & 0.2468 & 0.1294 & 0.1922 & 0.1339 & 0.3013 & 0.3975 & 0.2864 & 0.2774 \\
        \bottomrule
    \end{tabular}
    }
\end{table*}

\begin{table*}[h]
    \centering
    \caption{Detailed ROUGE-N F1 Scores across Topics.}
    \label{tab:topics_full}
    \setlength{\tabcolsep}{1.5pt}
    \resizebox{\textwidth}{!}{
    \begin{tabular}{lcccccccccccccccccccccccc}
        \toprule
        \textbf{Method} & \textbf{Health} & \textbf{Pol.} & \textbf{Hist.} & \textbf{Ind.} & \textbf{Edu} & \textbf{Fash.} & \textbf{Ent.} & \textbf{Home} & \textbf{Trsp.} & \textbf{Fin.} & \textbf{Lit.} & \textbf{Sci.} & \textbf{Crime} & \textbf{Sprt} & \textbf{S.Dev} & \textbf{Soc.} & \textbf{Game} & \textbf{Soft} & \textbf{Rel.} & \textbf{HW} & \textbf{Art} & \textbf{Trvl} & \textbf{Food} & \textbf{Adult} \\
        \midrule
        Dripper\_fallback & \textbf{0.9181} & \textbf{0.9321} & \textbf{0.8859} & \textbf{0.8945} & \textbf{0.9036} & \textbf{0.8025} & \textbf{0.8926} & \textbf{0.8571} & \textbf{0.8765} & \textbf{0.9121} & \textbf{0.8718} & \textbf{0.9008} & \textbf{0.9113} & \textbf{0.8950} & \textbf{0.9112} & \textbf{0.9208} & \textbf{0.8392} & \textbf{0.8884} & \textbf{0.9220} & \textbf{0.8861} & \textbf{0.8090} & \textbf{0.8630} & \textbf{0.8855} & \textbf{0.8445} \\
        Dripper & 0.8961 & 0.9161 & 0.8588 & 0.8768 & 0.8849 & 0.7889 & 0.8748 & 0.8539 & 0.8654 & 0.8905 & 0.8418 & 0.8790 & 0.8864 & 0.8774 & 0.8948 & 0.8982 & 0.8024 & 0.8736 & 0.8959 & 0.8548 & 0.7987 & 0.8558 & 0.8739 & 0.8347 \\
        magic-html & 0.7863 & 0.8196 & 0.7006 & 0.6945 & 0.7130 & 0.6099 & 0.6679 & 0.6187 & 0.6826 & 0.7639 & 0.6955 & 0.7499 & 0.8161 & 0.7036 & 0.7323 & 0.7134 & 0.6890 & 0.6613 & 0.7776 & 0.6762 & 0.5965 & 0.6712 & 0.6669 & 0.6396 \\
        readability & 0.7306 & 0.7956 & 0.6486 & 0.6294 & 0.6479 & 0.4481 & 0.6506 & 0.5180 & 0.5754 & 0.7007 & 0.6367 & 0.6924 & 0.7575 & 0.6490 & 0.6669 & 0.6342 & 0.6203 & 0.6125 & 0.7286 & 0.6037 & 0.5298 & 0.6582 & 0.5996 & 0.6118 \\
        trafilatura-html & 0.7183 & 0.7745 & 0.6426 & 0.6036 & 0.6400 & 0.4987 & 0.6370 & 0.5430 & 0.5921 & 0.6584 & 0.6673 & 0.6673 & 0.7497 & 0.6137 & 0.5733 & 0.6753 & 0.5972 & 0.6191 & 0.6762 & 0.5885 & 0.5068 & 0.6379 & 0.6396 & 0.6066 \\
        resiliparse & 0.7095 & 0.7447 & 0.6254 & 0.5651 & 0.6348 & 0.5120 & 0.6054 & 0.5616 & 0.6131 & 0.6110 & 0.6665 & 0.6665 & 0.7439 & 0.6085 & 0.6076 & 0.6183 & 0.5830 & 0.5913 & 0.6684 & 0.5608 & 0.5258 & 0.6514 & 0.6025 & 0.5724 \\
        trafilatura-md & 0.7028 & 0.7691 & 0.6283 & 0.5852 & 0.6273 & 0.5023 & 0.6266 & 0.5288 & 0.5827 & 0.6288 & 0.7245 & 0.6537 & 0.7362 & 0.6112 & 0.5306 & 0.6725 & 0.5858 & 0.6020 & 0.6665 & 0.5684 & 0.5034 & 0.6269 & 0.6511 & 0.6003 \\
        trafilatura-text & 0.6838 & 0.7452 & 0.6165 & 0.5704 & 0.6059 & 0.4776 & 0.6098 & 0.5086 & 0.5612 & 0.6104 & 0.7041 & 0.6369 & 0.7179 & 0.5961 & 0.5157 & 0.6528 & 0.5734 & 0.5806 & 0.6465 & 0.5532 & 0.4854 & 0.6061 & 0.6175 & 0.5828 \\
        html2text-md & 0.6231 & 0.6110 & 0.6215 & 0.5493 & 0.6174 & 0.3926 & 0.5893 & 0.4904 & 0.5603 & 0.6154 & 0.7196 & 0.6954 & 0.6154 & 0.6082 & 0.7345 & 0.6010 & 0.6164 & 0.6710 & 0.6758 & 0.5255 & 0.4853 & 0.5320 & 0.5152 & 0.5748 \\
        boilerpy3-text & 0.6444 & 0.6816 & 0.5245 & 0.5262 & 0.5422 & 0.4217 & 0.5237 & 0.5015 & 0.5550 & 0.4917 & 0.6578 & 0.5457 & 0.6616 & 0.5327 & 0.3947 & 0.5610 & 0.4754 & 0.4862 & 0.6242 & 0.4820 & 0.5027 & 0.5730 & 0.5717 & 0.5572 \\
        gne & 0.6181 & 0.6576 & 0.5112 & 0.5019 & 0.5338 & 0.3779 & 0.5127 & 0.4146 & 0.4802 & 0.4942 & 0.6860 & 0.5353 & 0.6571 & 0.5103 & 0.4722 & 0.5359 & 0.4449 & 0.4457 & 0.5425 & 0.4464 & 0.4347 & 0.4726 & 0.4695 & 0.5502 \\
        newsplease & 0.5951 & 0.5641 & 0.4556 & 0.4147 & 0.5087 & 0.3662 & 0.4866 & 0.4633 & 0.5024 & 0.4967 & 0.5131 & 0.4714 & 0.6414 & 0.5269 & 0.4523 & 0.4814 & 0.4331 & 0.4757 & 0.6131 & 0.4498 & 0.4822 & 0.4970 & 0.5120 & 0.4771 \\
        boilerpy3-html & 0.5618 & 0.5977 & 0.4645 & 0.4298 & 0.4961 & 0.3029 & 0.4646 & 0.3823 & 0.4502 & 0.4501 & 0.6459 & 0.5216 & 0.5655 & 0.4593 & 0.4284 & 0.5039 & 0.4422 & 0.4868 & 0.5619 & 0.3721 & 0.4069 & 0.4606 & 0.4646 & 0.5136 \\
        justtext & 0.5665 & 0.5212 & 0.4314 & 0.3967 & 0.4805 & 0.4042 & 0.4625 & 0.5041 & 0.5040 & 0.4804 & 0.4444 & 0.4320 & 0.6075 & 0.4916 & 0.3153 & 0.4459 & 0.4704 & 0.4270 & 0.5481 & 0.4625 & 0.4625 & 0.5471 & 0.5227 & 0.4681 \\
        goose3 & 0.5227 & 0.4992 & 0.4082 & 0.3844 & 0.4318 & 0.3604 & 0.4298 & 0.4284 & 0.4623 & 0.3881 & 0.3598 & 0.3975 & 0.5735 & 0.4677 & 0.3010 & 0.4151 & 0.3783 & 0.3829 & 0.5663 & 0.3977 & 0.3803 & 0.4492 & 0.4803 & 0.4108 \\
        readerlm & 0.2706 & 0.2981 & 0.2102 & 0.1940 & 0.2592 & 0.1157 & 0.1960 & 0.1671 & 0.2115 & 0.1956 & 0.2852 & 0.2538 & 0.3105 & 0.2260 & 0.1991 & 0.2383 & 0.2075 & 0.2043 & 0.2696 & 0.1998 & 0.1813 & 0.2015 & 0.2186 & 0.2419 \\
        \bottomrule
    \end{tabular}%
    }
\end{table*}

\begin{table*}[h]
    \centering
    \caption{Average ROUGE-N F1 Scores and Rankings across Topics and Formats. Dripper and its fallback variant consistently outperform all baselines.}
    \label{tab:domain_summary}
    \small
    \begin{minipage}[t]{0.48\textwidth}
        \centering
        \textbf{(a) Topic Classification}
        \resizebox{\textwidth}{!}{
        \begin{tabular}{lcc}
            \toprule
            \textbf{Method} & \textbf{Avg Score} & \textbf{Rank} \\
            \midrule
            Dripper\_fallback & 0.8843 & 1 \\
            Dripper & 0.8656 & 2 \\
            magic-html & 0.7019 & 3 \\
            readability & 0.6394 & 4 \\
            trafilatura-html-md & 0.6305 & 5 \\
            trafilatura-md & 0.6189 & 6 \\
            resiliparse & 0.6180 & 7 \\
            trafilatura-text & 0.5998 & 8 \\
            html2text-md & 0.5898 & 9 \\
            boilerpy3-text & 0.5397 & 10 \\
            gne & 0.5068 & 11 \\
            newsplease & 0.4966 & 12 \\
            justtext & 0.4783 & 13 \\
            boilerpy3-html-md & 0.4714 & 14 \\
            goose3 & 0.4334 & 15 \\
            readerlm & 0.2222 & 16 \\
            \bottomrule
        \end{tabular}}
    \end{minipage}
    \hfill

    \begin{minipage}[t]{0.48\textwidth}
        \centering
        \textbf{(b) Format Classification}
        \resizebox{\textwidth}{!}{
        \begin{tabular}{lcc}
            \toprule
            \textbf{Method} & \textbf{Avg Score} & \textbf{Rank} \\
            \midrule
            Dripper\_fallback & 0.8997 & 1 \\
            Dripper & 0.8778 & 2 \\
            magic-html & 0.7259 & 3 \\
            readability & 0.6875 & 4 \\
            trafilatura-html-md & 0.6769 & 5 \\
            resiliparse & 0.6731 & 6 \\
            trafilatura-md & 0.6640 & 7 \\
            trafilatura-text & 0.6455 & 8 \\
            html2text-md & 0.6247 & 9 \\
            boilerpy3-text & 0.5834 & 10 \\
            newsplease & 0.5610 & 11 \\
            gne & 0.5523 & 12 \\
            justtext & 0.5414 & 13 \\
            boilerpy3-html-md & 0.5261 & 14 \\
            goose3 & 0.4782 & 15 \\
            readerlm & 0.2515 & 16 \\
            \bottomrule
        \end{tabular}}
    \end{minipage}
\end{table*}

\end{document}